\pdfoutput=1

\documentclass[11pt]{article}

\usepackage {EMNLP2023}

\usepackage{times}
\usepackage{latexsym}

\usepackage[T1]{fontenc}

\usepackage[utf8]{inputenc}

\usepackage{microtype}

\usepackage{inconsolata}

\usepackage{graphicx}
\usepackage{caption}
\usepackage{subcaption}
\usepackage{enumitem}
\usepackage{mathtools}
\usepackage{amsthm}
\usepackage{amsmath}
\usepackage{amssymb}
\usepackage{amsfonts} 
\usepackage{nicefrac}       %
\usepackage{mathtools}
\usepackage{booktabs} %
\usepackage{multirow}
\usepackage{tcolorbox}
\usepackage{hyperref}
\usepackage{titlesec}
\usepackage{lipsum}%

\titlespacing{\paragraph}{%
  0pt}{%
  0.2\baselineskip}{%
  0.5em}%

\newtcbox{\hlprimarytab}{on line, rounded corners, box align=base, colback=green!10,colframe=white,size=fbox,arc=3pt, before upper=\strut, top=-2pt, bottom=-4pt, left=-2pt, right=-2pt, boxrule=0pt}
\newtcbox{\hlsecondarytab}{on line, box align=base, colback=red!10,colframe=white,size=fbox,arc=3pt, before upper=\strut, top=-2pt, bottom=-4pt, left=-2pt, right=-2pt, boxrule=0pt}
\newcommand{\dashifted}{\raisebox{0.5\depth}{\tiny$\downarrow$}}
\newcommand{\uashifted}{\raisebox{0.5\depth}{\tiny$\uparrow$}}
\newcommand{\da}[1]{{\tiny\hlprimarytab{\dashifted{#1}}}}
\newcommand{\ua}[1]{{\tiny\hlsecondarytab{\uashifted{#1}}}}

\newcommand\revision[1]{\textcolor{black}{#1}}

\newcommand\tf[1]{\textbf{#1}}

\title{Adapting Language Models to Compress Contexts}

\author{Alexis Chevalier$^{*}$ \quad Alexander Wettig$^{*}$\quad Anirudh Ajith \quad Danqi Chen \\
\textnormal{Department of Computer Science \& Princeton Language and Intelligence} \\
\textnormal{Princeton University} \\
\texttt{\{achevalier, anirudh.ajith\}@princeton.edu} \\ 
\texttt{\{awettig, danqic\}@cs.princeton.edu}\\ 
}

\begin{document}
\maketitle
\renewcommand{\thefootnote}{\fnsymbol{footnote}}
\footnotetext[1]{AC and AW contributed equally. This work was done when AC was at the Institute for Advanced Study and visited the Princeton NLP group.}
\renewcommand{\thefootnote}{\arabic{footnote}}

\begin{abstract}
{
Transformer-based language models (LMs) are powerful and widely-applicable tools, but their usefulness is constrained by a finite context window and the expensive computational cost of processing long text documents.
We propose to adapt pre-trained LMs into \mbox{\emph{AutoCompressors}}.
These language models are capable of compressing long contexts into compact \emph{summary vectors}, which are then accessible to the model as soft prompts.
Summary vectors are trained with an unsupervised objective,
whereby long documents are processed in segments, and summary vectors from all previous segments are used in language modeling. 
We fine-tune OPT and Llama-2 models on sequences of up to 30,720 tokens
and show that AutoCompressors can utilize long contexts to improve perplexity.
We evaluate AutoCompressors on in-context learning by compressing task demonstrations and find that summary vectors are good substitutes for plain-text demonstrations, increasing accuracy while reducing inference costs. 
Finally, we explore the benefits of pre-computing summary vectors for large corpora by applying summary vectors to retrieval-augmented language modeling and a passage re-ranking task.
Overall, AutoCompressors emerge as a simple and inexpensive solution to extend the context window of LMs while speeding up inference over long contexts.\footnote{Our code and models are publicly available at \href{https://github.com/princeton-nlp/AutoCompressors}{https://github.com/princeton-nlp/AutoCompressors}.}}

\end{abstract}

\section{Introduction}

\begin{figure}[t]
    \centering  
    \includegraphics[width=\linewidth]{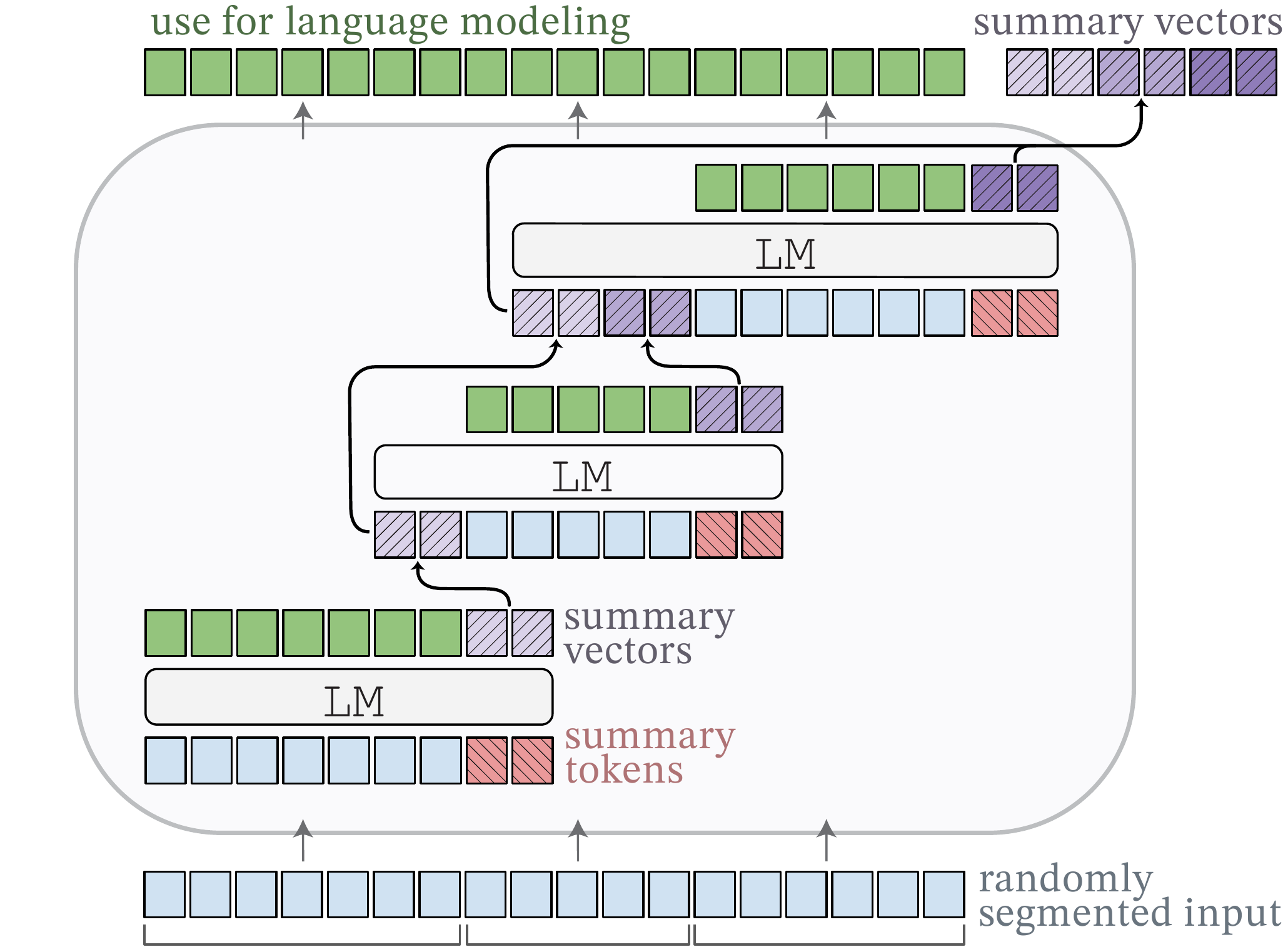}
    \caption{\textit{AutoCompressors} process long documents by recursively generating summary vectors which are passed as soft prompts to all subsequent segments.}  \label{fig:architecture}
\end{figure}

Transformer-based~\cite{vaswani} language models (LMs) have recently seen a sharp rise in popularity and are now receiving millions of queries, processing billions of tokens, and generating text for a wide variety of applications~\cite{gpt3,touvron2023llama,opt}. With this rise in popularity comes the challenge for researchers to make LMs more \emph{efficient}, to speed up inference and to deploy LMs at scale, while increasing their \emph{versatility}, thus allowing users to process more data in new ways.

With these goals in mind, we propose to teach pre-trained LMs the ability to compress text into \emph{summary vectors}.  Summary vectors are short soft prompts \cite{lester-etal-2021-power}, one or two orders of magnitude shorter than the pre-compressed plain text, that are obtained from the output states of a language model. Summary vectors serve two general purposes: they can help extend the language model's context window to very long documents with minimal computational overhead, and they help speed up inference on text for which summary vectors have been pre-computed and cached.

Our models, which we call AutoCompressors, are trained with a simple unsupervised learning objective that encourages the model to store essential information in the summary vectors.
Summary vectors are produced segment by segment from long documents and are used to improve language modeling in future segments (Figure \ref{fig:architecture}).
Our work builds on the recently proposed RMT architecture \cite{RMT} with a crucial difference:
we introduce \emph{summary accumulation}, in which summary vectors from all segments are concatenated to produce the summary of the entire document. We also train AutoCompressors with randomly segmented inputs so they can  better compress contexts of variable lengths in downstream tasks. 
We show that these innovations improve long-range information retention and enable new ways of reasoning over multiple passages.

\revision{
AutoCompressors can be initialized with pre-trained LMs to produce powerful and versatile models.
We fine-tune AutoCompressors from OPT-2.7B \cite{opt} and Llama-2-7B \cite{touvron2023llama} models on sequences from 6,144 up to 30,720 tokens with a single NVIDIA A100 GPU of 80GB memory. 
 We show that summary vectors are effective for improving perplexity over long documents and that these compression capabilities are robust to domain generalization. 
Our analysis suggests that AutoCompressors are able to reason over summary vectors, making them useful for a diverse set of downstream applications.}

We apply AutoCompressors to in-context learning (ICL) by compressing up to 90 in-context demonstrations. We consider 11 classification tasks, including 7 SuperGLUE tasks \cite{wang2019superglue}, and we find that summary vectors outperform few-shot ICL with a comparable number of in-context tokens on 8 out of 11 tasks.

 Finally, we explore two applications where \mbox{AutoCompressors} can reduce inference costs by pre-computing
summary vectors for large corpora. 
First, we adopt a setting for retrieval-augmented language modeling \cite{shi2023replug}. 
We find that for equal sequence lengths, using summary vectors achieves 1.5$\times$ the perplexity gains compared to plain-text passages, and outperforms retrieval-augmented methods for similar computational budgets.
Secondly, we consider a zero-shot passage re-ranking task \cite{sachan-etal-2022-improving}.
We establish that re-ranking passages based on their summary vectors
achieves the best trade-off between re-ranking performance and inference throughput.

In summary, our main contributions are the following: 
(1) We introduce a method for extending LMs to long context windows under small-scale computational requirements
by learning to generate summary vectors. We propose summary accumulation and training with randomized segmenting as key features of AutoCompressors. 
(2) We show that summary vectors encode useful information for downstream tasks and can be used to reduce the inference cost of in-context learning.
(3) We demonstrate the benefits of pre-computing summary vectors for large corpora and using AutoCompressors in conjunction with retrievers.

\section{Related Work}\label{section:background}
\paragraph{Soft prompts}
Soft prompt tuning is an effective method to adapt pre-trained Transformers without updating existing parameters \cite{lester-etal-2021-power, zhong-etal-2021-factual, liu-etal-2022-p}.
Newly initialized embeddings are prepended to the input sequence (the ``soft prompt''),
and optimization is performed with respect to these new parameters while the rest of the model is frozen.
It is one of many parameter-efficient fine-tuning methods \cite{lialin2023scaling} and is related to prefix tuning, where newly initialized parameters are prepended to the attention states instead \cite{prefixtuning}.

\paragraph{Prompt compression}
\citet{promptcompression}
propose to learn a soft prompt $\sigma$ to compress the information contained in a context $x$.
Given a pre-trained language model $p_\text{LM}$, they draw continuations $y \sim p_\text{LM}(\cdot \mid x)$ based on $x$ and use a distillation objective to align the model's predictions conditioned on the soft prompt $p_\text{LM}(y \mid \sigma)$ to the predictions conditioned on the context $p_\text{LM}(y \mid x)$.
\citet{promptcompression} find that soft prompts retain high-level information and facilitate controllable generation.
However, the approach requires running the optimization for every new context $x$, with no knowledge transfer between similar contexts.
In contrast, our \mbox{AutoCompressors} learn to predict their own soft prompts $\sigma$ as a function of $x$.

\paragraph{Context distillation}
A related line of work \cite{askell2021general, snell2022learning} aims to distill in-context information, e.g., instructions, into an unprompted student model.
In concurrent work, \citet{gisting} teach models to compress instructions into short key-value attention prefixes. Our approach differs by learning to compress any context information, including long documents, and results in more compact soft prompts.

\paragraph{Long-range Transformers}
A number of architectural modifications have been proposed to 
scale Transformers to longer context lengths while reducing the high memory costs of full attention.
These include restricting and sparsifying the attention window \cite{TransformerXL, child2019generating}, approximating the attention 
\cite{Rae2020Compressive, zheng2022linear, choromanski2021rethinking},
as well as introducing recurrent elements \cite{mega, RMT}, conditional computation \cite{colt5}, 
and retrieving previous tokens from the context at the output layer \cite{zhong-etal-2022-training}.
See \citet{tay2023survey} for a comprehensive survey of efficient long-range architectures.

Most of these architectures typically require expensive training from scratch, or will deviate substantially from a pre-trained initialization.\footnote{In our pre-liminary experiments, even fine-tuning a pre-trained OPT-2.7b model with Transformer-XL-style training \cite{TransformerXL} caused optimization difficulties and deterioriated the pre-trained model quality.} 
Moreover, many language models lack the inductive bias to extrapolate to longer sequences \cite{press2022train}.
While AutoCompressors could in principle be trained from scratch, we show that they offer a straightforward solution for extending the context window of pre-trained models to longer sequences.

\section{Method}
\label{sec:arch}

We describe how we adapt a pre-trained language model to compress text into summary vectors. An overview of our architecture is shown in Figure \ref{fig:architecture}.

\paragraph{Summary vectors} 
The AutoCompressor builds on the RMT architecture \citep{RMT}.
We extend the input vocabulary of the base model by $\kappa$ special summary tokens $\texttt{<Sum>}_i$
and initialize $\kappa$ new \mbox{input} embeddings.\footnote{
When fine-tuning OPT models, 
we observe benefits with initializing the embeddings of the summary tokens with the pre-trained embedding for the end-of-sequence token $\texttt{</s>}$.}
When we append the sequence $\texttt{<Sum>}_1 \dots \texttt{<Sum>}_\kappa$ to an input, it signals to the model to output special \textit{summary vectors} of the preceding context.
These vectors can then be passed to the next text segment as a soft prompt of \mbox{length $\kappa$}.
Since the embedding spaces of pre-trained language models can span thousands of dimensions,
we expect that this mechanism has a high capacity for passing information to subsequent segments.
Furthermore, a soft prompt can interpolate between many token embeddings, and therefore represent more abstract concepts than a single discrete token \cite{promptcompression}.

\paragraph{Summary accumulation}
We split long documents into segments $S_1, \ldots, S_n$
and process them sequentially.
\citet{RMT} incorporate information from previous segments by prepending the compressed summary $\sigma_{i-1}$ produced from $S_{i-1}$ to the embedded inputs of $S_i$.
We propose \textit{summary accumulation}, which allows for a direct information pathway between each segment and all segments preceding it: 
we concatenate the summary vectors $\sigma_{1}\ldots, \sigma_{i-1}$ to form $\sigma_{<i}$ and prepend $\sigma_{<i}$ to $S_i$.
Note that the length of $\sigma_{<i}$ is now $(i-1)\kappa$, which grows linearly with the document length.

\paragraph{Positional embeddings}
\revision{
When using a base Transformer architecture with absolute positional embeddings, such as the OPT architecture \cite{opt}, we do not add positional embeddings to the summary tokens $\texttt{<Sum>}_i$, nor to the summary vectors.
This allows us to use all pre-trained position embeddings as context tokens and makes it possible to scale the model to an arbitrary number of compression steps during training. The model still preserves the order of summary tokens due to their separate token embeddings.}

\revision{If the base Transformer uses relative positional embeddings, such as RoPE \cite{su2022roformer}, we apply the positional embedding to the summary tokens and vectors without any further modification.}

\subsection{Training Summary Vectors}\label{subsection:training}
We use a simple unsupervised training approach which encourages the model to learn to compress contexts over multiple steps.

\paragraph{Training objective}\label{sec:training}
Write $(x_{1}^i, \dots, x_{m_i}^i)$ for the segment $S_i$ for every $i \leq n$, where $m_i$ is the number of tokens in $S_i$. Conditioning on the concatenated summary vectors $\sigma_{<i}$, we project the Transformer outputs with the language modeling head 
to obtain the next-token probabilities $p(x_{t}^i \mid x_1^i, \ldots, x_{t-1}^i, \sigma_{<i})$.
We minimize the cross-entropy loss over the entire document:
$$
\mathcal{L} = -\frac{1}{N} \sum_{i=1}^n \sum_{t=1}^{m_i} \log p(x_{t}^i \mid x_1^i, \dots, x_{t-1}^i, \sigma_{<i}).
$$
where $N$ is the total number of tokens. This objective retains the pre-trained language model's abilities on the first segment $S_1$ and it incentivizes the model to store useful information in the summary vectors,
which future segments can leverage to make better token predictions.

Unlike \citet{promptcompression}, we do not train with a knowledge distillation objective, 
since the pre-trained LM has a limited context window as a teacher, whereas the AutoCompressor student learns to process much longer documents.

\paragraph{Randomized segmenting}
We randomly vary the lengths $m_i$ of the segments $S_i$ during training, subject to the condition that each segment fits into the model's context window.
This allows AutoCompressors to compress documents of different lengths and improves performance under evaluation with fixed-length segments (see Figure \ref{fig:ablations}).

\paragraph{BPTT with stop-gradients}
We employ backpropagation through time (BPTT) and gradient checkpointing \cite{gradient_checkpointing} for each segment to reduce the size of the computational graph.
In addition, we compute and cache summary vectors and stop their gradients after 2 compression steps,
similar to caching past attention states in Transformer-XL training \citep{TransformerXL}.
This assumes that for learning to compress the useful information in $S_i$,
it is sufficient to predict the tokens in the adjacent $S_{i+1}$.
In Figure \ref{fig:ablations}, we confirm that this incurs no penalty when predicting long segments, 
while further reducing GPU memory requirements.

\begin{table*}[ht]
    \centering
    \resizebox{\textwidth}{!}{
        \begin{tabular}{lcccccccccc}
            \toprule
                &\multicolumn{5}{c}{\tf{In-domain}} & \multicolumn{5}{c}{\tf{Out-of-domain}} \\
                \cmidrule(lr){2-6} \cmidrule(lr){7-11}
            \multicolumn{1}{r}{\textit{Segments}} & \multicolumn{3}{c}{\textit{----------- \; 1 \; -----------}} & \textit{-- 2 --} & \textit{-- 3 --} & \multicolumn{3}{c} {\textit{----------- \; 1 \; -----------}} & \textit{-- 2 --} &\textit{-- 3 --}\\
             \multicolumn{1}{r}{Context tokens} & 128 & 512 & 2048 & 4096  & 6144  & 128 & 512 & 2048 & 4096 & 6144 \\
            \midrule
                Extended FA$^\dagger$    & {6.33}$^\dagger$\ua{1.0\%} & {6.15}$^\dagger$\da{2.1\%} & {5.94}$^\dagger$\da{5.4\%} & -    & -    & {8.57}$^\dagger$\ua{0.5\%} & {8.28}$^\dagger$\da{2.9\%} & {7.93}$^\dagger$\da{7.0\%} & -    & -    \\
            \midrule
                RMT       & 6.42\ua{2.2\%} & 6.19\da{1.4\%} & 6.02\da{4.1\%} & 6.02\da{4.1\%} & {6.01}\da{4.3\%}  & 8.76\ua{2.7\%} & 8.44\da{1.1\%} & 8.21\da{3.8\%} & 8.20\da{3.9\%} & {8.20}\da{3.9\%}  \\
                AutoCompressor    & 6.14\da{2.2\%} & 6.04\da{3.8\%} & 5.98\da{4.8\%} & 5.94\da{5.4\%} & \textbf{5.93}\da{5.6\%} & 8.39\da{1.6\%} & 8.26\da{3.2\%} & 8.17\da{4.2\%} & 8.12\da{4.8\%} & \textbf{8.10}\da{5.0\%}  \\
              \bottomrule
        \end{tabular}
    }
    \caption{Held-out perplexity on 2,048 tokens, while varying the length of the preceding context (all the experiments are based on OPT-2.7B models). For RMT and AutoCompressor, we condition on summary vectors. We also report the perplexity gains compared to the fine-tuned OPT baseline without extra context, which achieves 6.28 in-domain and 8.53 out-of-domain (gains shown in colored numbers).    $\dagger$: Although the extended full attention (Extended FA) achieves similar or slightly better perplexity, it uses up to 2,048 additional tokens and cannot extend further. However, the AutoCompressor uses only $50 \times 3=150$ summary vectors to process 6,144 context tokens.
    }
    \label{table:ppl}
\end{table*}

\begin{table}[t]
    \centering   
    \resizebox{\linewidth}{!}{
      \begin{tabular}{l ccc  c}
      \toprule
            \multicolumn{1}{r}{\textit{Segments}} & \textit{-- 0 --}& \textit{-- 7 --} & \textit{-- 14 --} & \multicolumn{1}{|c}{CUDA} \\ 
        \multicolumn{1}{r}{Context tokens} & 0 & 14336 & 28672 & \multicolumn{1}{|c}{memory} \\ 
      \midrule
      RMT-1.3B & 13.18 & 12.50 & 12.50 & \multicolumn{1}{|c}{54GB} \\
      AutoCompressor-1.3B & 13.21 & 12.49 & \textbf{12.47} & \multicolumn{1}{|c}{38GB} \\
      \midrule
      RMT-2.7B & - & - & - & \multicolumn{1}{|c}{\texttt{OOM}} \\
      AutoCompressor-2.7B & 11.86  & 11.21 & \textbf{11.18} & \multicolumn{1}{|c}{75GB}  \\
      \bottomrule
    \end{tabular}
   }
   
   \caption{Evaluation results for AutoCompressors  trained on sequences of 30,720 tokens and evaluated on Books3 (in-domain) and Gutenberg (out-of-domain). 
 We train with a single NVIDIA A100 GPU and report the CUDA memory required for fine-tuning using a single sequence per batch. AutoCompressors require less memory  because we stop gradients after two segments.} \label{tab:very_long_range}
\end{table}

\begin{table}[t]
    \centering   
    \resizebox{\linewidth}{!}{
      \begin{tabular}{lcccccc}
      \toprule
        \multicolumn{1}{r}{\textit{Segments}} &  \textit{-- 0 --}  & \multicolumn{3}{c}{\textit{ \textendash\textendash\textendash\textendash\textendash\textendash\textendash{} 1 \textendash\textendash\textendash\textendash\textendash\textendash}} & \textit{-- 2 --} & \textit{-- 3 --} \\ 
        \multicolumn{1}{r}{Context tokens}  & 0 & 128 & 512  & 2048 & 4096 & 6144 \\ 
      \midrule
      Llama-2        &  5.52 & 5.30  & 5.15 & 4.98 & - & -  \\
      Extended FA     &   5.40     &  5.19  & 5.06 &  4.88 & 4.80 & 4.76  \\
      \midrule
      AutoCompressor     &  5.40 & 5.23  & 5.16 & 5.11  & 5.08 & \textbf{5.07}  \\
      \bottomrule
    \end{tabular}
   }
   
   \caption{\revision{Evaluation results for our AutoCompressor trained from Llama-2 7B on sequences of 6,144 tokens. For the AutoCompressor, we condition on summary vectors. For Llama-2 and the Extended Full Attention (Extended FA), we condition on plain text tokens.}} \label{tab:llama2_ppl}
\end{table}

\section{Language Modeling Evaluation}

\revision{
In this section, we train AutoCompressors and evaluate their long-range language modeling capabilities  by sampling long sequences which we split into segments of 2,048 tokens. We fix the final segment and compress the previous $n$ segments. We track the perplexity of the final segment when conditioning on the summary vectors for each $n$.}

\revision{
We conduct our main experiments and ablations with OPT models \cite{opt} of 1.3B or 2.7B parameters, fine-tuned on 2B tokens from the Pile \cite{pile}.
In Section \ref{sec:language_modeling}, we evaluate an AutoCompressor on sequences of 8,000 tokens and compare to an equivalent RMT model and an Extended Full Attention baseline. In Section \ref{sec:very_long_range}, we fine-tune an AutoCompressor on sequences of 30,000 tokens to demonstrate the feasibility on very long sequences.
Finally, in Section \ref{sec:llama2_ppl}, we scale up AutoCompressors by fine-tuning a Llama-2-7B model on 15B tokens from RedPajama \cite{together2023redpajama}.
Full model hyperparameters and data information can be found in Appendix \ref{appendix:hyperparams}.}

\subsection{Experiments on 8K-Token Sequences}
\label{sec:language_modeling}

\paragraph{Setting} 
\revision{We initialize all models with the 2.7B-parameter OPT model and fine-tune on 2B tokens from 4 domains form the Pile \citep{pile}.}
Our AutoCompressor uses $\kappa=50$ summary tokens and is fine-tuned with summary accumulation over four segments, each ranging from 1,024 to 2,048 tokens. Compressing 2,048 tokens into $50$ summary vectors achieves a compression rate  of 40 tokens per summary vector. 
We use the following baselines:
 
\begin{enumerate}[leftmargin=0.03\textwidth,noitemsep,nolistsep]
\item
We fine-tune an OPT-2.7B baseline on our data. This model is limited to sequences of 2,048 tokens due to pre-training.
\item
Extended full attention:  We fine-tune OPT-2.7B on sequences of up to 4,096 tokens by extending the model's positional embeddings. 
We initialize the embeddings for positions $[2049..4096]$ with the embeddings for positions $[1..2048]$. We are not able to extend the context beyond 4,096 tokens due to GPU memory limitations.
\item
RMT-2.7B: We fine-tune an RMT model on our data with $\kappa=50$ summary vectors.
\end{enumerate}

\revision{We evaluate on documents of 8,192 tokens, drawn from the 4 training domains or 4 held-out domains.}
We generate summary vectors for up to 3 segments of 2,048 tokens, but also for single segments as short as 128 tokens.  
For the extended full-attention baseline we prepend the previous context tokens to the context window.

\paragraph{Results} We show the results in Table \ref{table:ppl}. 
We find that the AutoCompressor benefits from long contexts of 6,144 tokens and consistently outperforms the RMT model.  

We also find that the AutoCompressor benefits from much shorter sequences than seen during training,
unlike RMT. See also Figure \ref{fig:ablations} and Table \ref{table:no_context} for the usefulness of randomized segmenting.

While extended full attention performs the best on 4,096-long sequences, 
we observe a trade-off for shorter contexts where AutoCompressors achieve the best performance. 
We also stress that the \mbox{AutoCompressor} attends to at most 150 additional soft prompts during evaluation, whereas the full attention model is given an additional 2,048 tokens.

These trends hold for both in-domain and out-of-domain evaluation. 
However, the gap between the AutoCompressor and the full-attention baseline increases in the out-of-domain setting, suggesting that the summary vectors generalize slightly less than pre-trained attention heads.

\subsection{Experiments on 30K-Token Sequences}
\label{sec:very_long_range}

\paragraph{Setting}
\revision{We fine-tune OPT-1.3B and OPT-2.7B as AutoCompressors on 2B tokens but train on sequences of 30,720 tokens with 20 compression steps.\footnote{Due to the scarcity of very long sequences in the Pile, we only train on data from the Books3 domain, and use the Gutenberg domain as out-of-domain evaluation.}}
We use 50 summary tokens, randomized segmenting, and stop-gradients as before. We also fine-tune an RMT model from OPT-1.3B, to use as a baseline. We are not able to fine-tune a 2.7B-parameter RMT baseline because the RMT method leads to an out-of-memory error.

All models are evaluated on the final 2,048 held-out tokens of documents of size 30,720 tokens by compressing all previous 2,048-token segments.

\paragraph{Results} We collect our results in Table \ref{tab:very_long_range}. The evaluation shows that both AutoCompressor models learn to utilize the entire 28K tokens to reduce perplexity, while the RMT baseline does not benefit from doubling the number of context tokens from 14K to 28K.
This shows that summary accumulation effectively captures long-range dependencies in documents. We also report the CUDA memory requirements for fine-tuning each model in Table \ref{tab:very_long_range}. We train with one NVIDIA A100 GPU with 80GB of memory. Stopping gradients reduces CUDA memory and makes it possible to fine-tune an AutoCompressor from OPT-2.7B, while fine-tuning with RMT leads to out-of-memory at that scale.

\subsection{Scaling Up AutoCompressors to Llama-2}\label{sec:llama2_ppl}
\paragraph{Setting} We fine-tune a 7B-parameter Llama-2 model as an AutoCompressor on a single GPU by freezing the model and optimizing only the summary token embeddings and the attention weights via LoRA \cite{hu2022lora}.
The model is trained on 15B tokens from RedPajama \cite{together2023redpajama}, split into sequences of 6,144 tokens, and we use 50 summary tokens, randomized segmenting, and stop-gradients. We also fine-tune an Extended Full Attention baseline on the same dataset. The context window of the pre-trained model is extended by increasing the $\theta$ value in RoPE following \cite{codellama}.

We compare both models to the pre-trained Llama-2-7B model, which has a context window of 4,096 tokens. All models are evaluated on the final 2,048 tokens of 8,192-token documents.
 
\paragraph{Results} We collect our results in Table \ref{tab:llama2_ppl}. The \mbox{AutoCompressor} benefits from the entire context to reduce perplexity: compressing a 4,096-token context into 100 summary vectors achieves similar perplexity to the Extended Full Attention baseline with 512 plain text tokens, and compressing a 6,144-token context into 150 summary vectors further improves perplexity slightly.  Moreover, we find that summary vectors preserve perplexity when short contexts are compressed.

However, Llama-2 and the Extended Full Attention baseline outperform the \mbox{Auto}Compressor when longer contexts are provided. Further research is needed to construct summary vectors that preserve all of the context information.

\subsection{Analysis}

\begin{figure}[t]
    \centering  
    \includegraphics[width=\linewidth]{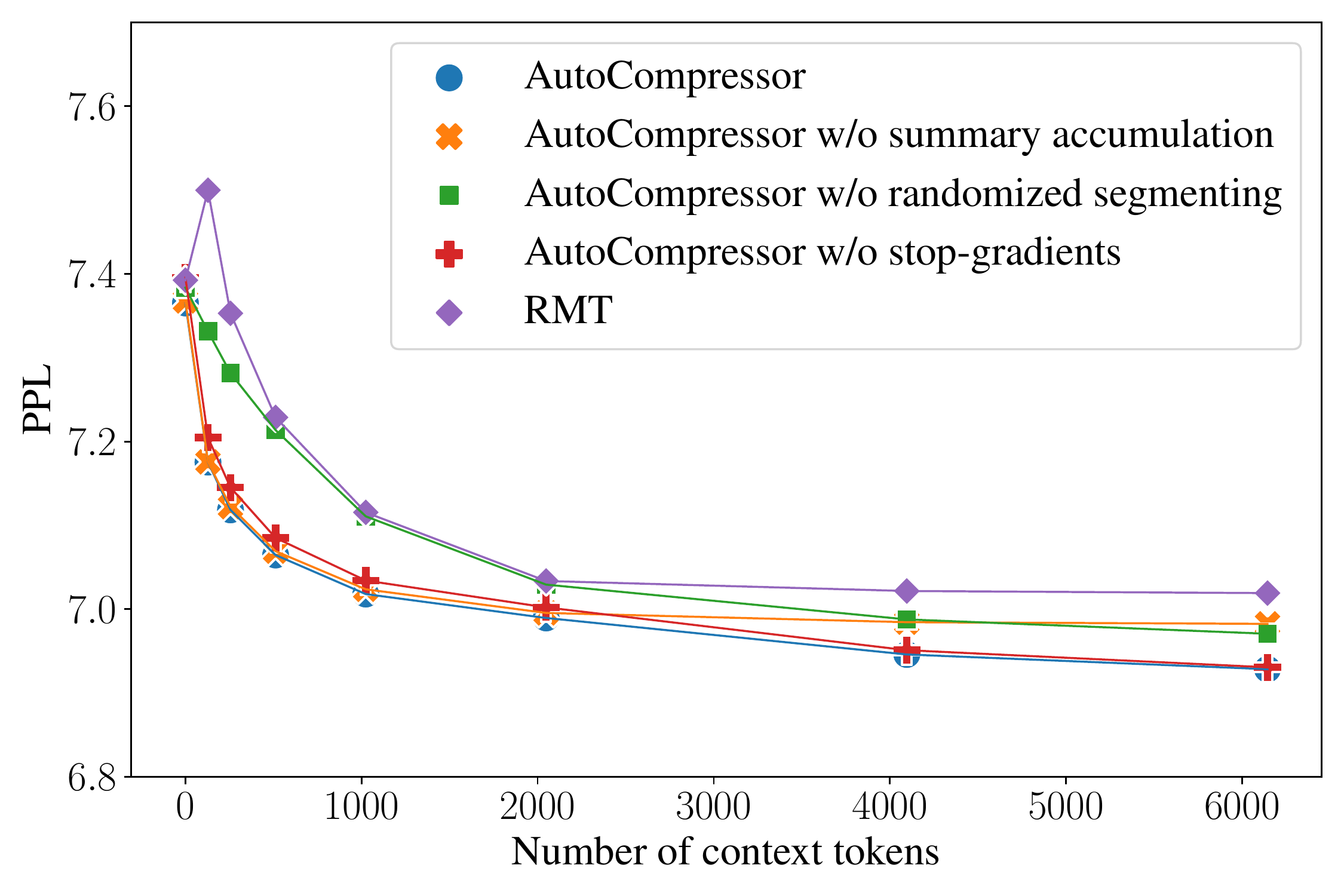}
    \caption{Perplexity on 2048 held-out tokens given different numbers of compressed tokens. Compression is performed on up to 3 segments of 2048 tokens. Ablations show that the different components of our fine-tuning strategy help boost performance and that stop-gradients do not impact performance.   }\label{fig:ablations}
\end{figure}

\paragraph{Ablations} 

We  train OPT-2.7B models without randomized segmenting, summary accumulation, or stop gradients. The results are shown in Figure \ref{fig:ablations}.
We find that randomized segmenting leads to better compression of short segments, but still improves perplexity when compressing multiple 2048 token segments. As expected, summary accumulation helps improve perplexity beyond one compressed segment. We also confirm that stopping gradients does not impact performance despite reducing GPU memory requirements. In Table \ref{tab:very_long_range}, we also show that stopping gradients helps reduce GPU memory.

We also train AutoCompressors with $\kappa =20 $, $50$, $70$ or $100$ summary tokens and report the held-out perplexity results in Table \ref{tab:summary_tokens} in the Appendix. 
Surprisingly, we find that performance does not increase with longer soft prompts, and $\kappa = 50$ performs the best overall. We hypothesize that learning a larger number of summary vectors may require a larger training budget.

\paragraph{Token-level analysis}
We seek to better understand how summary vectors benefit individual token predictions. In Figure \ref{fig:tokenwise_analysis} in the Appendix, we show perplexity gains at each token position for the AutoCompressor with summary vectors and for the extended full-attention baseline. 

We find that conditioning on summary vectors improves perplexity over  all 2048 token positions.
We observe that the extended full attention baseline outperforms the AutoCompressor   at the start of the sequence,
whereas the AutoCompressor achieves the best performance towards the end of the sequence. 
This shows that summary vectors effectively capture  long-range textual dependencies.

In Appendix \ref{appendix:performance_analysis}, we show  examples of sentences and tokens which benefit the most from summary vectors. We find that summary vectors contain salient information, such as names or dates, and that the model can reason over summary vectors. This confirms that summary vectors are useful summaries of the compressed text.

\section{Compressing Demonstrations for In-Context Learning}

\begin{table*}[t]
    \centering
    \resizebox{\textwidth}{!}{
\begin{tabular}{rcccccccccccc}
    \toprule
    & \textbf{AG News} & \textbf{SST-2} & \textbf{BoolQ} & \textbf{WIC} & \textbf{WSC} & \textbf{RTE} & \textbf{CB} & \textbf{COPA} & \textbf{MultiRC} & \textbf{MR} & \textbf{Subj} \\
    \midrule
Zero-shot & $63.3_{(0.0)}$ & $67.7_{(0.0)}$ & $67.4_{(0.0)}$ & $50.8_{(0.0)}$ & $43.3_{(0.0)}$ & $58.8_{(0.0)}$ & $42.9_{(0.0)}$ & $52.5_{(0.0)}$ & $52.5_{(0.0)}$ & $57.4_{(0.0)}$ & $49.3_{(0.0)}$ \\
    \midrule
    50 summary vecs. & $79.6_{(4.9)}$ & $\textbf{94.2}_{(1.6)}$ & $\textbf{70.1}_{(3.3)}$ & $51.6_{(2.1)}$ & $47.7_{(8.7)}$ & $66.3_{(7.0)}$ & $46.4_{(18.7)}$ & $84.5_{(1.0)}$ & $52.6_{(2.8)}$ & $91.5_{(1.0)}$ & $53.5_{(3.6)}$ \\
    100 summary vecs. & $\textbf{87.6}_{(1.2)}$ & $92.6_{(3.3)}$ & $66.3_{(2.8)}$ & $52.5_{(2.2)}$ & $42.9_{(2.5)}$ & $63.5_{(6.6)}$ & $\textbf{64.5}_{(5.9)}$ & $85.9_{(0.4)}$ & $\textbf{56.1}_{(1.2)}$ & $90.7_{(2.6)}$ & $57.0_{(5.6)}$ \\
    150 summary vecs. & $85.4_{(3.4)}$ & $92.3_{(2.9)}$ & $68.0_{(1.8)}$ & $\textbf{52.8}_{(1.5)}$ & $49.9_{(7.6)}$ & $65.3_{(6.6)}$ & $54.8_{(5.8)}$ & $\textbf{86.1}_{(0.6)}$ & $54.8_{(2.2)}$ & $91.1_{(2.2)}$ & $56.6_{(7.9)}$ \\
    \midrule
    ICL (150 tokens) & $74.5_{(2.2)}$ & $92.4_{(3.1)}$ & $67.4_{(0.0)}$ & $52.4_{(2.7)}$ & $\textbf{51.8}_{(6.9)}$ & $69.1_{(2.1)}$ & $46.4_{(23.0)}$ & $80.0_{(1.9)}$ & $52.5_{(0.0)}$ & $79.7_{(15.7)}$ & $57.9_{(10.7)}$ \\
    ICL (750 tokens) & $81.2_{(4.1)}$ & $93.8_{(1.2)}$ & $67.7_{(2.7)}$ & $52.4_{(2.0)}$ & $40.0_{(5.7)}$ & $\textbf{73.1}_{(3.5)}$ & $50.3_{(2.8)}$ & $82.6_{(1.6)}$ & $47.0_{(3.2)}$ & $\textbf{91.6}_{(0.8)}$ & $\textbf{60.7}_{(14.8)}$ \\
 \bottomrule
\end{tabular}
    }
        \caption{\revision{Evaluation of the ICL performance of the Llama-2 7B model. Each summary is 50 tokens-long and corresponds to a segment of 750 tokens' worth of demonstrations. 
        We also report accuracies when prompting the AutoCompressor with 150 and 750 tokens' worth of plaintext  demonstrations as baselines. Note that for BoolQ and MultiRC, demonstrations are too long to fit into 150 tokens.}}
    \label{tab:icl_results_llama}
\end{table*}

In this section, we study the usefulness of summary vectors for performing downstream tasks. We show that in-context demonstrations can reliably be compressed down into summary vectors to improve performance while also increasing efficiency on a diverse set of NLP benchmarks.

\paragraph{Evaluation}
\revision{
We evaluate the in-context learning abilities of the AutoCompressor based on Llama-2-7B from Section \ref{sec:llama2_ppl} on eleven classification and multiple-choice question-answering datasets.
For each dataset, we evaluate the effect of compressing 1, 2 or 3 segments
of demonstrations into 50, 100 or 150 summary vectors. For each segment, we include as many demonstrations as possible until we reach 750  tokens. For SST-2, this corresponds to 30 demonstrations per segment on average. We compare this compression approach with the results obtained by prompting the model using 150 and 750 tokens' worth of plain-text demonstrations.}

\revision{
We use contextual calibration \cite{pmlr-v139-zhao21c} and class-balanced sampling when these techniques improve performance on a validation set. For each dataset, we report the mean accuracy and standard deviation  over 7 random seeds. The detailed settings for each dataset can be found in Table \ref{tab:icl_datasets}. In Table \ref{tab:expanded_icl_results} in the Appendix, we also compare the ICL performance of our OPT-2.7B based AutoCompressor models against the RMT baseline and a pre-trained OPT-2.7B, and include the performance of the pre-trained Llama-2-7B model.}

\paragraph{Results} 
\revision{We show evaluation results in Table \ref{tab:icl_results_llama}.  Results show that summary vectors consistently improve performance over the zero-shot baseline. Furthermore, summary vectors increase accuracy compared to 150 tokens worth of plain demonstrations on 8/11 tasks. 
On 8 tasks (AG News, \mbox{SST-2}, BoolQ, WiC, WSC, CB, COPA and MultiRC), summary vectors also out-perform ICL with 750 tokens' worth of plain text demonstrations.
Summary vectors emerge as a strong alternative to plain text demonstrations, as they increase accuracy while reducing inference cost.}

\revision{In Table \ref{tab:expanded_icl_results} (Appendix \ref{app:icl}), we find that the OPT-2.7B AutoCompressor achieves higher accuracies than the RMT baseline on 8 out of 11 tasks and that the RMT model does not benefit from multiple compression steps. This shows that summary accumulation is an effective mechanism for compressing in-context demonstrations. We also observe that our fine-tuned Llama-2 AutoCompressor has substantially worse zero-shot accuracy on some tasks compared to the Llama-2 initialization, and slightly worse ICL performance. We suspect that this is due to domain mismatch in our fine-tuning data and the Llama-2 pre-training corpus.}

\section{Compressing Retrieval Corpora for Efficient Inference}

We study the usefulness of pre-computing summary vectors for large collections of documents. These can be stored and later retrieved for efficient inference. 
Since inference is typically more expensive than storage, this approach has the potential to achieve good practical trade-offs.

\subsection{Retrieval-augmented Language Modeling}\label{sec:replug}
\begin{figure}[t]
    \centering  
    \includegraphics[width=\linewidth]{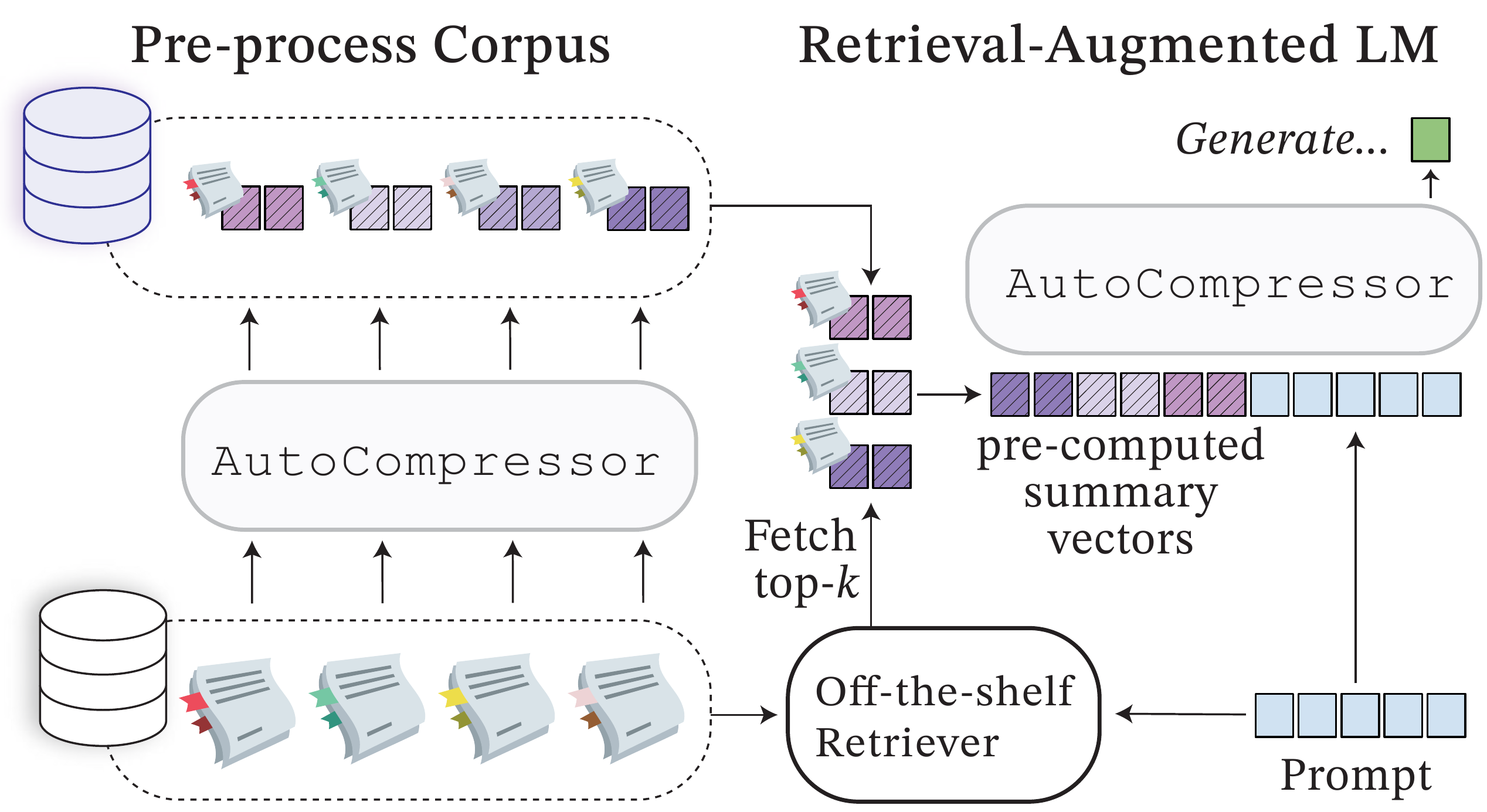}
    \caption{Efficient retrieval-augmented language modeling with AutoCompressors.
    Large corpora can be pre-processed into compressed summary vectors which can be stored cheaply. 
    Upon retrieval, compressed summaries are fused for efficient access to multiple documents in a single forward pass.}
    \label{fig:retrieval}
\end{figure}

\begin{table*}[t!]
    \centering   
    \resizebox{\linewidth}{!}{
    \begin{tabular}{ll cccc cccc}
    \toprule
                                    &  & \multicolumn{4}{c}{\tf{Perplexity Gain} (\%)} & \multicolumn{4}{c}{\tf{Throughput} (examples/s)}    \\
    \cmidrule(lr){3-6}  \cmidrule(lr){7-10}
     Passages            &                       & top-$1$ & top-$2$  & top-$5$ & top-$10$ & top-$1$ & top-$2$ & top-$5$ & top-$10$ \\
    \midrule    
    50 tokens            & REPLUG                & -0.64  & 0.58 & 1.68 & 2.35 & 51 & 38     & 16    & 9     \\
    50 tokens            & Fused Passages        & 0.71 & 1.01  & 1.70 & 2.60 & 28 & 27     & 23     & 17   \\
    512 tokens $\to$ 50 sum. vecs.   &     Fused Summaries     &  \tf{1.04} & \tf{1.67} & \tf{2.63}  & \tf{3.74} & 28  & 27  & 23    & 17\\
    \midrule
    512 tokens & REPLUG         & -1.47  & 2.24  & 5.25  & 8.30     & 18  & 10  & 6 & 3\\
    \bottomrule
\end{tabular}

    } 
    \caption{ 
        PPL gains  ($\%$) from different retrieval-augmented language modeling settings, over the no-retrieval baseline. We evaluate the OPT-2.7B AutoCompressor and  we  report throughput on a single NVIDIA A100 GPU for each method without batching examples.  Fused Summaries outperforms Fused Passages and REPLUG with 50-token passages. Moreover, Fused Summaries top-10 outperforms  REPLUG top-2 with 512-token passages while also gaining a 1.7$\times$ throughput increase.
    }
    \label{tab:replug_results}
\end{table*}

Retrieval-augmented language models improve token predictions by retrieving information from a data store.
A number of approaches have been proposed to infuse external knowledge in the input layer \cite{guu2020retrieval, shi2023replug}, intermediate layers \cite{retro} or at the output layer \cite{Khandelwal2020Generalization, zhong-etal-2022-training}.  

\paragraph{REPLUG}  Our case study focuses on REPLUG \cite{shi2023replug}, which is a simple method for combining a pre-trained language model with an off-the-shelf retriever to improve language modeling performance.
Given access to an external corpus $\mathcal{C}$, REPLUG retrieves $k$ passages $\mathcal{D}=\{d_1, \ldots, d_k \}$ based on a segment $x$ to score the next segment $y$. The overall probability for $y$ is computed by ensembling the predictions based on different passages:
\begin{equation*}
    p(y \mid  x, \mathcal{D}) = \sum_{d \in \mathcal{D}} \lambda(d, x) \cdot p(y \mid \textsc{Concat}(d, x)),
\end{equation*}
where $\lambda(d,x)$ are the normalized similarity scores from the retriever
and $\textsc{Concat}(d, x)$ denotes concatenation of $p$ and $x$. This method incurs a substantial overhead, since it requires $k$ forward passes over sequences $\textsc{Concat}(d, x, y)$.

\paragraph{Fused Summaries}
We introduce a  setting for retrieval-augmented language modeling close to fusion-in-decoder \citep{izacard-grave-2021-leveraging}. We concatenate the summary vectors of retrieved passages $\mathcal{D}$ to form the \emph{ fused summary \mbox{vectors}},
$\sigma_\mathcal{D} = \textsc{Concat}(\sigma_{d_k}, \ldots, \sigma_{d_1})$, where $d_k, \ldots, d_1$ are ordered from least-to-most relevant.
This resembles summary accumulation as described in Section \ref{sec:arch}.
We also find it useful to smooth probability scores and re-order the retrieved passages based on their summary vectors (Appendix \ref{app:replug}).
Figure \ref{fig:retrieval} gives an overview of our \mbox{approach}.

\paragraph{Fused Passages} 
We establish a baseline for fusing summary vectors by concatenating the plain-text passages
 and computing smoothed probabilities, see Appendix \ref{app:replug}.  Unlike summary vectors, this method is limited by the model's context window.

\paragraph{Experiments}

We evaluate the  OPT-2.7B AutoCompressor introduced in Section \ref{sec:language_modeling} without any additional fine-tuning.
Similar to \citet{shi2023replug}, we retrieve from the Pile. We use Books3, FreeLaw, GitHub, Wikipedia, Gutenberg, ArXiv, HackerNews, and YoutubeSubtitles. We index 10B tokens for each domain, which are split into passages of 512  or 50 tokens. 

We sample segments of 256 tokens from the Pile validation data, using the first 128 tokens as context $x$ for retrieval and the last 128 tokens $y$ for evaluation. We use the Contriever model \citep{izacard2022unsupervised} for retrieval, and retrieve the top 10 passages. 
We also deduplicate our data by removing passages that overlap with $x$ by 64 tokens.

\paragraph{Results}
Results are shown in Table \ref{tab:replug_results}. We find that Fused Summaries outperforms Fused Passages and REPLUG when 50-token passages are retrieved. We measure throughput empirically and show that for 10 retrieved documents,  Fused Summary Vectors remains inexpensive.  We note that compressing the 10B token datasets results in disk space of 5TB  per domain when stored in half-precision format.\footnote{For comparison, storing the transformer output at every single token (e.g., in an encoder-decoder setting) would take up 51 TB, and storing all attention states would be 3,276 TB.} Therefore Fused Summaries achieves a good trade-off between storage costs and throughput.

Moreover, Fused Summaries outperforms REPLUG top-2 with 512-token passages and sees a 1.7x throughput increase, which shows that the model benefits from the diversity of compressed documents. However, REPLUG top-10 outperforms Fused Summaries. We leave it as future work to explore how to produce higher quality summary vectors to better utilize the compressed passages.

We note that fusing summary vectors is effective despite a mismatch in training since we draw independent summary vectors from separate documents.
Furthermore, our AutoCompressor model is only ever trained to accumulate 3 sets of summary vectors, and yet it benefits from fusing the summary vectors of up to 10 documents.

\subsection{Unsupervised Passage Re-ranking}\label{subsection:reranking}

Finally, we consider the case study of passage re-ranking, in which a fast off-the-shelf retriever like BM25 retrieves a large set of candidate passages, and a more capable re-ranker refines the ranking to increase the rank of the most relevant passages.

\paragraph{Method}
\citet{sachan-etal-2022-improving} introduce an effective method for leveraging language models as re-rankers with no additional supervision or fine-tuning.
Given a query $q$ and a set of candidate passages $\{p_1, \ldots, p_k\}$, the language model scores the likelihood of the query $q$ conditioned on the prompt \texttt{``Passage: \{$p_i$\}. Please write a question based on this passage.''} for each passage $p_i$ and re-ranks the passages based on the scores.

\paragraph{Experiments}
We consider the task of re-ranking BM25 passages on the NQ test set \cite{balachandran-etal-2021-investigating}
and compare out-of-the-box AutoCompressors with 20 and 50 summary tokens to pre-trained OPT models from 125M to 2.7B parameters.\
We pre-compute summary vectors for 21M passages from a Wikipedia corpus \cite{karpukhin-etal-2020-dense}, which requires 2.1TB and 5.4TB disk space in half precision for 20 and 50 summary vectors respectively. 
We measure the quality of the re-ranked results using Recall@20.

\begin{figure}[t]
    \centering
    \includegraphics[width=\linewidth]{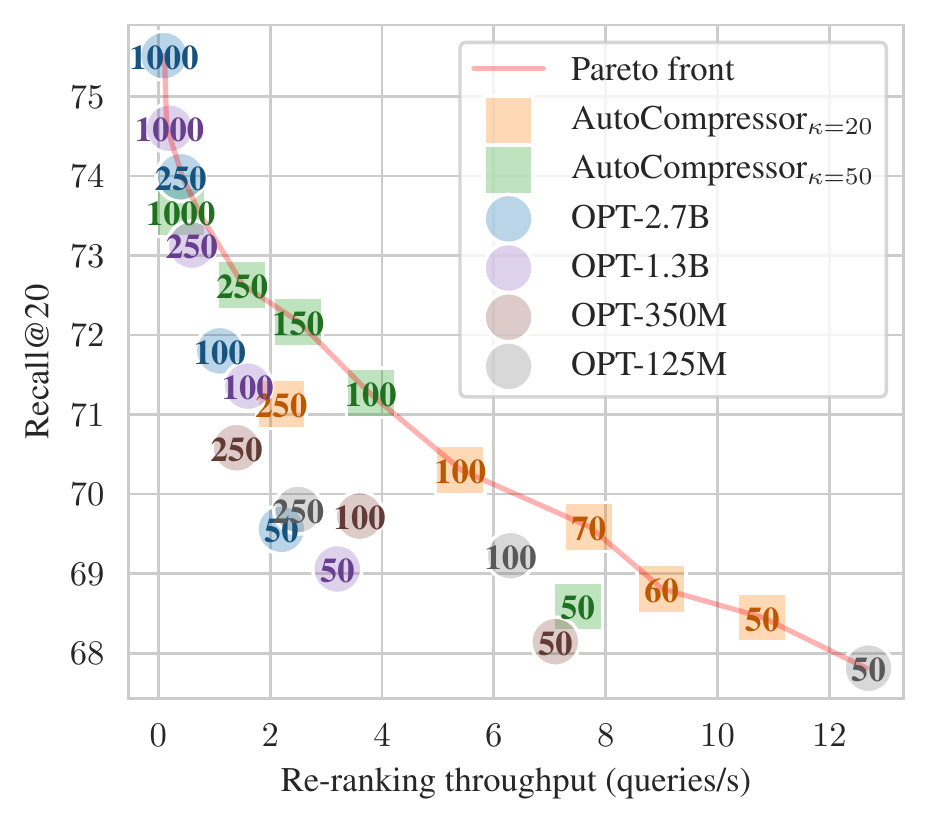}
    \caption{We compare AutoCompressors (squares) in an unsupervised passage re-ranking setting
    to pre-trained language models (circles).
    The number on each data point shows how many passages retrieved by BM25 are re-ranked, and the vertical axis shows the Recall@20 performance of the re-ranking system on the NQ test set.
    We consider the throughput on a single NVIDIA A100 GPU and assume that multiple queries cannot be batched.
    By leveraging pre-computed summary vectors for passages,
    AutoCompressors lead to re-ranking solutions that lie on the Pareto front of recall vs. compute.}
    \label{fig:recall}
\end{figure}

\paragraph{Results} 
The results are shown in Figure \ref{fig:recall}.
We measure throughput for individual un-batched queries on a single NVIDIA A100 80GB GPU and assume that the latency of loading summary vectors is negligible.
Although the passages are only 100 words long, resulting in low compression rates,
summary vectors substantially speed up the inference, while sacrificing on performance less than smaller models.
This leads to a Pareto-optimal trade-off between compute and performance and demonstrates
that summary vectors often retain sufficient information from a passage to assess its relevance for a particular query.

\section{Conclusion}
We have introduced a training strategy for adapting pre-trained LMs into AutoCompressors,
which recursively compress contexts into summary vectors.
Our experiments indicate that summary vectors retain important contextual information, that they can encode in-context demonstrations, and that they can be used in retrieval settings. 
Summary vectors can also be pre-computed, cached and re-used. This offers practical efficiency gains by reducing the size of the attention window.
Significant future work remains in scaling AutoCompressors to bigger models and improving the quality of summary vectors to further close the gap with full attention over long-range contexts.

\section*{Limitations}
\begin{enumerate}[leftmargin=0.03\textwidth,noitemsep,nolistsep]
    \item
    We only apply AutoCompressors to OPT models of up to 2.7B parameters and a Llama model of 7B parameters.
    Future work needs to establish how AutoCompressors perform for even larger models. As the summary vector dimension grows, there is promise for retaining more information per vector.

    \item Our results suggest that summary vectors ignore some useful information that is accessible via full attention. Additionally, models do not always benefit from increasing the number of summary vectors.
    We suspect that the training signal for learning summary vectors efficiently might be limited by pre-trained models being very good at making predictions from the plain-text tokens in the current segment. Future work is needed to improve this optimization.

    \item Summary accumulation still leads to quadratic complexity with increasing number of segments, albeit at a much lower rate than full attention. Future work may explore ways to combine many summary vectors more efficiently.
\end{enumerate}

\section*{Acknowledgments}
We thank Mengzhou Xia, Howard Chen, Vishvak Murahari, Aatmik Gupta, Zirui Wang, Jiatong Yu,  and the members of the Princeton NLP group for helpful discussion and valuable feedback. This research is supported by an NSF CAREER award (IIS-2239290), a Sloan Research Fellowship, and a Data Science Research Award from Adobe. 
AC also gratefully acknowledges support from the Minerva Research Foundation.

\bibliography{anthology,custom}
\bibliographystyle{acl_natbib}

\appendix

\clearpage

\section{Models and Data}\label{appendix:hyperparams}

All models are fine-tuned from OPT models on the Pile. We conduct our experiments using a single NVIDIA A100 80GB GPU and we use Flash Attention \cite{dao2022flashattention} as an efficient implementation of exact attention over long sequences. We also use gradient checkpointing between compressed segments to reduce GPU memory. 

\subsection{OPT Experiments on 8K Tokens}
We fine-tune our models on 2B tokens from the Pile. We sample 500M tokens from the following Pile subdomains: Books3, FreeLaw, GitHub and Wikipedia. 

The following  models use a learning rate of 2e-5, a batch size of 130K tokens, 1,00 warm-up steps, and the Adam optimizer \cite{Adam}:

\begin{enumerate}[leftmargin=0.03\textwidth,noitemsep,nolistsep]
    \item The fine-tuned OPT-2.7B baseline is fine-tuned on documents of up to 2,048 tokens. 
    \item The extended full-attention baseline is fine-tuned on documents of up to 4,096 tokens by extending the positional embeddings of OPT-2.7B to 4,096 positions.
    We initialize the embeddings for positions $[2049..4096]$ with the embeddings for positions $[1..2048]$.
    \item The RMT baseline is fine-tuned on documents of up to 8,192 tokens. Each document is segmented into four segments of 2,048 tokens. We use $\kappa = 50$ summary vectors but we do not use summary accumulation, randomized segmenting, or stop-gradients.
    \item Our AutoCompressor is fine-tuned on documents of up to 6,144 tokens. Each document is randomly segmented  into four segments such that the first two segments add up to 3,072 tokens. The length of each segments ranges from 1,024 to 2,048 tokens. We use $\kappa=50$ summary vectors and summary accumulation. We stop gradients every two compression steps.
\end{enumerate}

\medskip
All models are evaluated on documents sampled from the Pile with a fixed length of 8,192 tokens. We sample 610 documents from each of the following domains: Books3, FreeLaw, GitHub, Wikipedia (in-domain), and ArXiv, Gutenberg, HackerNews, YoutubeSubtitles (out-of-domain). Examples of documents from each of those domains can be found in Tables \ref{tab:examples_in} and \ref{tab:examples_out}.

\subsection{OPT Experiments on 30K Tokens}

We fine-tune our models on 2 billion tokens from the Books3 subdomain of the Pile. All models are fine-tuned on documents of up to 30,720 tokens. We use a learning rate of 2e-5, a batch size of 130k tokens, 1,000 warm-up steps and the Adam optimizer. 

\begin{enumerate}[leftmargin=0.03\textwidth,noitemsep,nolistsep]
    \item RMT-1.3B uses $\kappa = 50$ summary vectors and is fine-tuned without summary accumulation, randomized segmenting, or stop-gradients. Each document is split into 15 segments of 2,048 tokens Even with gradient checkpointing, attempting to fine-tune a 2.7B parameter RMT model on this dataset leads to an out-of-memory error.
    
    \item The AutoCompressor models are fine-tuned from OPT-1.3B and 2.7B on documents of up to 30,720 tokens. Each document is split into 20 segments such that 
    segment $2i$ and segment $2i+1$ add up to 3,072 tokens. The length of each segment is randomly sampled between 1,024 and 2,048. 
    We use $\kappa = 50$ summary vectors with summary accumulation and we stop gradients every two compression steps.
\end{enumerate}

\medskip
All models are evaluated on documents of 30,720 tokens from the Pile. We use 1,000 documents from Books3 (in-domain) and 1,000 documents from Gutenberg (out-of-domain).

\subsection{Llama-2 Experiments on 8K Tokens}

We fine-tune our Llama-2 models on 15B tokens from RedPajama. We sample 1B tokens from long documents in ArXiv, Books, C4, GitHub,  as well as 10B tokens from CommonCrawl, 800M from Wikipedia and 70M tokens from StackExchange.

Both our AutoCompressor and our Extended Full Attention baseline are fine-tuned from Llama-2-7B on sequences of 6,144 tokens with LoRA \cite{hu2022lora} parameter efficient fine-tuning applied to the attention heads. We use a LoRA dimension of 16 applied to the QKV- and Out-projections. We use a learning rate of 4e-4, a batch size of 200K tokens, 5,000 warm-up steps and the Adam optimizer. For the AutoCompressor, we also optimize the newly initialized summary token embeddings.

We train our AutoCompressor in the same way as the OPT-2.7B AutoCompressor, with $\kappa = 50$, randomly segmenting each sequence into four semgents, and stopping gradients every two compression steps. 
The Extended Full Attention baseline is fine-tuned with a RoPE $\theta$ value of 80,000. 

We evaluate our models on 500 sequences of 8,192 tokens from each of ArXiv, Books, C4, GitHub, StackExchange, and 5,000 sequences from CommonCrawl. 

\section{No-context Language Modeling}
 In Table \ref{table:no_context}, we verify that our fine-tuning strategy does not significantly affect the language modeling capabilities of the OPT AutoCompressors when no summary tokens are given. We find that the AutoCompressor performs slightly better than the RMT model and significantly better than the extended full attention model when no additional context is given. Moreover, the AutoCompressor almost matches the OPT02.7B fine-tuned baseline, with perplexity increasing by less than 1\%.
 
\begin{table}[h]
    \centering
    \resizebox{\linewidth}{!}{
    \begin{tabular}{lcc}
        \toprule
            & {In-domain} & {Out-of-domain} \\
        \midrule
            OPT-2.7B  & 7.53\ua{19.9\%} & 9.19\ua{7.7\%} \\
            OPT-2.7B fine-tuned  & 6.28 & 8.53 \\
        \cmidrule(lr){1-3}
            AutoCompressor-2.7B &  6.31\ua{0.5\%} & 8.60\ua{0.8\%} \\
            RMT-2.7B & 6.34\ua{1.0\%} & 8.62\ua{1.1\%} \\
            \midrule
            Extended full attention & 6.57\ua{6.4\%} & 8.94\ua{4.8\%} \\
        \bottomrule
    \end{tabular}
    }
    \caption{Held-out perplexity of all models on 2048 tokens without summary vectors or additional context. }
    \label{table:no_context}
\end{table}

\section{AutoCompressor Ablations}\label{app:ablations}

We train OPT AutoCompressor models as in Section \ref{sec:language_modeling} while varying $\kappa=20, 50, 70, 100$.  In Table \ref{tab:summary_tokens}, we report the perplexity evaluation on documents of 8192 tokens across all evaluation domains. 

\begin{table}[h]
    \centering   

    \resizebox{0.7\linewidth}{!}{
    \begin{tabular}{r cccc}
        \toprule
        & \multicolumn{4}{c}{Compressed tokens} \\
        \cmidrule{2-5}
        $\kappa$ & 0      & 2048      & 4096      & 6144     \\
        \midrule
        20             & \textbf{7.36}   & 7.05      & 7.01      & 7.00     \\
        50             & 7.37   & \textbf{6.99}      & \textbf{6.94}      & \textbf{6.93}     \\
        70             & 7.41   & 7.01      & 6.97      & 6.95     \\
        100            & 7.48   & 7.07      & 7.01      & 7.00     \\ 
        \bottomrule
    \end{tabular}
    }
    \caption{Held-out perplexity across all evaluation domains for AutoCompressors based on OPT-2.7B trained with different numbers of summary tokens $\kappa$. We observe that $\kappa=50$ performs the best overall.}
    \label{tab:summary_tokens}
\end{table}

\section{Token-level AutoCompressor Analysis}\label{appendix:performance_analysis}

\begin{figure}[h]
    \centering  
    \includegraphics[width=0.95\linewidth]{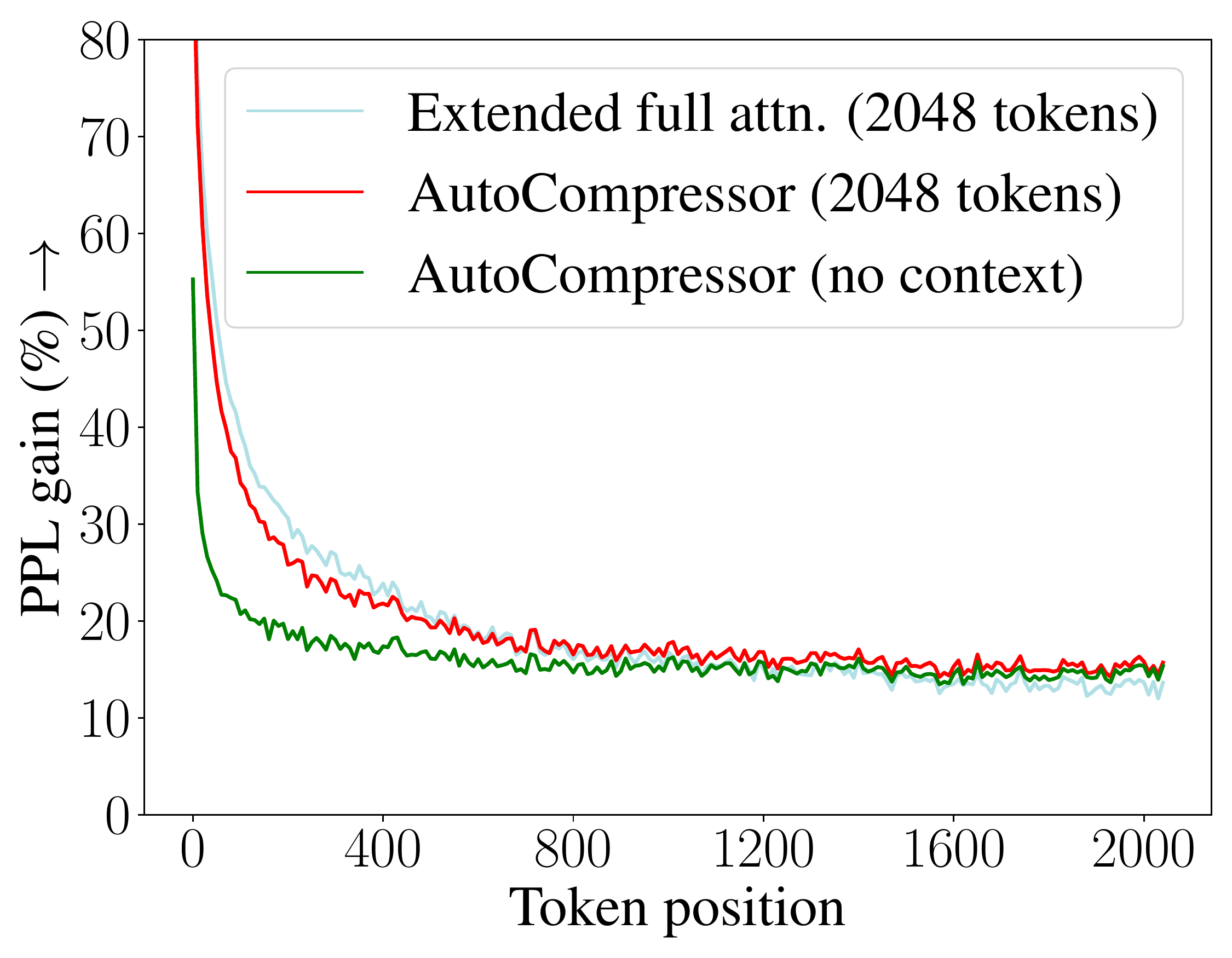}
    \caption{We plot the perplexity gain over OPT-2.7B for our AutoCompressor model and the 4096-extended attention baseline. We track the perplexity at each token position  in sequences of 2048 tokens.
    The AutoCompressor model almost matches the strong extended-attention baseline at the start of sequences and outperforms it at the end of sequences.    }     \label{fig:tokenwise_analysis}
\end{figure}

In Figure \ref{fig:tokenwise_analysis}, we plot the perplexity gains achieved by the OPT AutoCompressor  and the extended full attention baseline from Section \ref{sec:language_modeling} over the pre-trained OPT-2.7B model. We plot the gains achieved by the AutoCompressor both without any additional context and with the summary vectors obtained from 2048 compressed tokens.

Results show that the summary vectors help reduce perplexity over the entire 2,048-token segment. This shows that summary vectors do not only contain information which helps continue the previous sequence. 

Figure \ref{fig:tokenwise_analysis} also shows that the extended full-attention baseline benefits more from the additional 2,048 context tokens than the AutoCompressor at the start of the sequence, but that the AutoCompressor achieves stronger gains at the end of the sequence. This shows that summary vectors effectively capture  long-range textual dependencies and that fine-tuning AutoCompressors produces more robust models than fine-tuning extended full-attention models.

In Tables \ref{tab:examples_in} and \ref{tab:examples_out}, we give hand-picked examples of sequences from each evaluation domain, highlighting which tokens benefit the most from the compressed context. We compress the first 300 tokens in every document from the evaluation set and evaluate on the following 100 tokens. In the notation of Section \ref{sec:training}, we measure the perplexity gain of each token as \[
\frac{p(x^2_t \mid x^2_1, \ldots, x^2_{t-1}, \sigma_1)}{p(x^2_t \mid x^2_1, \ldots, x^2_{t-1})}.\]
For each example, we record the top 3-5 most improved token predictions.

We find that the tokens which benefit the most from the summary vectors are often interpretable. Names of characters, dates, and locations are often copied through the summary vectors (see the examples for Wikipedia, FreeLaw, or HackerNews). We also find that the model is able to reason over the summary vectors, as the tokens which benefit the most are sometimes not explicitly present in the compressed context, but are closely associated with the domain of speech (see the examples for Books3, Gutenberg and YoutubeSubtitles.). Finally, we find that summary vectors are often useful for continuing the previous sentence (see the GitHub example.)

\section{In-Context Learning Details}\label{app:icl}
We evaluate on in-context examples of the following datasets:
AG News (topic classification, \citet{zhang2015character_ag_news}), SST-2 (sentiment analysis, \citet{socher2013recursive_sst-2}), BoolQ (Boolean Questions, \citet{clark-etal-2019-boolq}), WiC (Word-in-Context, word sense dismabiguation, \citet{pilehvar-camacho-collados-2019-wic}), WSC (Winograd Schema Challenge, coreference resolution, \citet{levesque2012winograd}), RTE (Recognizing Textual Entailment, \citet{dagan2005pascal_rte1, bar2006second_rte2, bentivogli2009fifth_rte4}), CB (CommitmentBank,  \citet{demarneffe2019commitmentbank}), COPA (Choice of Plausible Alternatives, \citet{roemmele2011choice}), MultiRC (Multi-Sentence Reading Comprehension, \citet{khashabi-etal-2018-looking}), MR (Movie Reviews, \citet{MR}), Subj (Subjectivity, \citet{Subj}. 
We follow the GPT-3 prompt templates \cite{gpt3} and detail our evaluation setting for OPT and Llama-2 in Table \ref{tab:icl_datasets}.

In Table \ref{tab:expanded_icl_results}, we compile evaluation results for OPT-2.7B, Llama-2-7B, as well as our AutoCompressor and RMT models.

\section{Fused Retrieval-augmented Language Modeling}
\label{app:replug}

\begin{table}[h]
    \centering
    
    \resizebox{\linewidth}{!}{
    \begin{tabular}{l cccc}
        \toprule
                         & \multicolumn{4}{c}{Perplexity Gain (\%)}  \\
        \cmidrule(lr){2-5} 
        Passages                &  top-$1$ & top-$2$ & top-$5$ & top-$10$ \\
        \midrule    
             Fused Summaries             & 1.04 & \textbf{1.67} & \textbf{2.63} & \textbf{3.74}  \\
            Fused Summaries w/o re-ranking & 1.04 & 1.52 & 2.02 & 2.63 \\
        \midrule
    \end{tabular}
    }
    \caption{ PPL gains  ($\%$) over the no-retrieval baseline for Fused Summary with and without re-ranking. In re-ranking, we order the passages based on the $\ell_2$ norms of their summary vectors before concatenating the summary vectors, whereas w/o re-ranking we use the retrieval scores from the Contriever model. Re-ranking consistently produces higher perplexities.}
    \label{tab:fused_summary_ablation}
\end{table}

We provide details and ablations for our proposed REPLUG alternative. Inspired by fusion-in-decoder \citep{izacard-grave-2021-leveraging}, we fuse summary vectors or passages in a single forward pass.

\paragraph{Fused Summary Vectors}
The summary vectors of retrieved passages $\mathcal{D}$ are concatenated in order of increasing retrieval scores to form \emph{fused summary \mbox{vectors}},
$\sigma_\mathcal{D} = \text{Concat}[\sigma_{d_k}, \ldots, \sigma_{d_1}]$.
This resembles summary accumulation as described in Section \ref{sec:arch}, but differs in that the retrieved summary vectors were produced independently rather than recursively. Nevertheless, we find that AutoCompressors transfer well to this setting.

Furthermore, we find it beneficial to smooth the conditioned probabilities with the unconditioned probabilities $p(y \mid x)$, and compute
\[
    p(y \mid x, \mathcal{D}) = \frac{p(y \mid \text{Concat}[\sigma_\mathcal{D}, x]) + p(y\mid x)}{2}.
\]

We also show that language-modeling performance improves when 
$\mathcal{D}$ is re-ordered based on the smallest $\ell_2$ distance between the summary vectors $\{\sigma(d_1), \ldots, \sigma(d_k)\}$ and $\sigma_x$. This incurs negligible overhead since $\sigma_x$ can be constructed during the same forward pass which computes $p(y \mid x)$.
The ablation for this is shown in Table \ref{tab:fused_summary_ablation}

\paragraph{Fused Passages} 
We establish a baseline for Fusing Summary Vectors by concatenating the corresponding plain-text passages
\mbox{$D = \text{Concat}[d_k, \ldots, d_1]$} and computing
\[
p(y \mid x, \mathcal{D}) = \frac{p(y \mid \text{Concat}[D, x]) + p(y\mid x)}{2}.
\]

Note that this approach is quickly limited by the size of the pre-trained language model's context window, especially when retrieving many long passages.

\begin{table*}[t]
    \centering
    \resizebox{\textwidth}{!}{
    \begin{tabular}{r p{14cm} p{7cm} p{2cm}}
        \toprule
          \textbf{Domain} & \textbf{Compressed context} & \textbf{Evaluation sequence} & \textbf{Most \mbox{improved} \mbox{tokens}}\\
         \midrule
         Books3 & 
        Surrealism—not for Breton's depreciation of "Red Front," but for a seemingly insignificant aside. In early March, before sending the text to press, Breton showed it to {\color{cyan} Ar}{\color{violet} agon}. The latter consented to the publication, with one exception: a footnote in which Breton quoted the PCF official's remark (which {\color{cyan} Ar}{\color{violet} agon} had earlier reported to him) about "complicating the simple, healthy relations between men and women"—a clear illustration, Breton felt, of "just how much bad faith or mental indigence we were up against." {\color{cyan} Ar}{\color{violet} agon} considered internal Party statements to be confidential, and asked that the footnote be removed; according to him, Breton "spontaneously crossed out the note on the galleys with a delete mark that I can still recall... saying that he wanted to give the Party no excuse for expelling me." But when $\_$The {\color{blue} Poverty} of {\color{teal} Po}etry$\_$ came off press the next day, the incriminating footnote was still there.

        Whether Breton retained the note as a test of {\color{cyan} Ar}{\color{violet} agon}'s loyalty, or whether he deemed this example of PCF stupidity too good to waste, or whether the printer simply neglected to make the correction, no one has ever established. But the result was that this single act came to represent for {\color{cyan} Ar}{\color{violet} agon} every philosophical difference, stricture, and humiliation that had ever darkened his long friendship with Breton. On March 10, he responded to the tract via an anonymous note in

& $\_$L'{\color{magenta} Human}it\'e$\_$ : "Our comrade {\color{cyan} Ar}{\color{violet} agon} informs us that he has absolutely nothing to do with the publication of a pamphlet entitled $\_$The {\color{blue} Poverty} of {\color{teal} Po}etry$\_$... He wishes to make it clear that he entirely disavows both the contents of this pamphlet and the attention it has drawn to his name, every Communist being duty-bound to condemn the attacks contained in this pamphlet as incompatible with the class struggle." This short paragraph was the only notice he ever saw fit to give of
& {\color{blue} Poverty}

{\color{teal} Po}

{\color{cyan} Ar}

{\color{violet} agon}

{\color{magenta} Human}
\\
Wikipedia 
&   </s>{\color{magenta} Shi} {\color{blue} Ce}                 

{\color{magenta} Shi} {\color{blue} Ce} (; born 15 December 1985) is a Chinese deaf female table tennis player. She has represented China at the {\color{teal} De}aflympics four times from {\color{cyan} 2005}-2017. {\color{magenta} Shi} {\color{blue} Ce} has been regarded as one of the finest athletes to have represented China at the {\color{teal} De}aflympics, having won 14 medals at the event since making her debut in the {\color{cyan} 2005} {\color{violet} Summer} {\color{teal} De}aflympics.
                       
Biography     

{\color{magenta} Shi} {\color{blue} Ce} was born in Yichun, Heilongjiang on 15 December 1985. She was born with an ear condition that impaired her hearing which resulted in her deafness and has congenital malformation in her right ear. Her parents decided to consult a doctor and took her to an hospital in the Zhejiang Province in order to cure her ear impairment when she was just five years old. The doctor suggested
 that surgery would cause facial paralysis after {\color{magenta} Shi} {\color{blue} Ce}'s parents demanded for a surgery. {\color{magenta} Shi} {\color{blue} Ce} took the sport of Table tennis and started playing it at the age of nine.
     
Career         

{\color{magenta} Shi} {\color{blue} Ce} has won 14 medals in her {\color{teal} De}aflympic career as a Table tennis player including 11 gold medals. {\color{magenta} Shi} {\color{blue} Ce} was eligible to compete at the National Games of China despite her deafness, in 2015. In the competition, she secured gold medals in singles, doubles, mixed doubles and in the team events.                                                                                         
                                     
{\color{cyan} 2005} {\color{violet} Summer} {\color{teal} De}aflympics                                                                                                                                                                         
{\color{magenta} Shi} {\color{blue} Ce} made her first appearance at an international sports

&  event during the {\color{cyan} 2005} {\color{violet} Summer} {\color{teal} De}aflympics and excelled on her debut Deaflympic event after winning gold medals in the women's singles, doubles and in the mixed doubles. She was also the part of the Chinese Table tennis team which secured the silver medal in the 2005 Deaflympics. In the same year, she received the Deaf Sportswoman of the Year award from the ICSD for her remarkable performances at the 2005 Summer Deaflympics. {\color{magenta} Shi} {\color{blue} Ce}                  
& {\color{blue} Ce}

{\color{teal} De}

{\color{cyan} 2005}

{\color{violet} Summer}

{\color{magenta} Shi}\\
Github 
& </s> import sys

import datetime

def basic(arguments):

    \quad import api

    \quad critic = api.critic.startSession(for$\_$testing=True)
    
    \quad repository = api.repository.fetch(critic, name="critic")
    
    \quad branch = api.branch.fetch(critic, 
    
    \quad\quad\quad repository=repository, name=arguments.review)
        
    \quad review = api.review.fetch(critic, branch=branch)
    
    \quad alice = api.user.fetch(critic, name="alice")
    
    \quad bob = api.user.fetch(critic, name="bob")
    
    \quad dave = api.user.fetch(critic, name="dave")
    
    \quad erin = api.user.fetch(critic, name="erin")

    \quad all$\_$comments = api.comment.fetchAll(critic)
    
    \quad assert isinstance(all$\_$comments, list)

    \quad EXPECTED $ = {\color{cyan} \{}$
        {\color{blue} 0}: $\{$ "{\color{violet} text}": "This is a general {\color{magenta} issue}.",
             "{\color{teal} location}": None,
& 
    \quad "type": "{\color{magenta} issue}", "state": "open" {\color{cyan} $\}$},
        
     \quad   {\color{blue} 1} : $\{$"{\color{violet} text}": "This is a general note.",
     
     \quad\quad   "{\color{teal} location}": None,     
       
     \quad\quad    "type": "issue",
& {\color{blue} 1} 

{\color{teal} location}

{\color{cyan} $\}$}

{\color{violet} text}

{\color{magenta} issue}
 \\ 
FreeLaw 
& 8

By the end of 1975, Farmers National was insolvent and under investigation by the Florida Department of Insurance.  The Miami newspapers published a series of articles describing the relationship between Hauser and the company.  Lawrence Lee, an attorney for an Arizona union group, investigated Farmers National in connection with an {\color{teal} Old} {\color{blue} Security}-Farmers National proposal.  He was told by the Florida insurance department that Farmers National was connected with Hauser, that it had been injected with questionable {\color{magenta} assets} which were being investigated by the department, and that it had been fined \$5,000 for failing to disclose both Hauser's ownership and a loan to one of its directors.  Lee contacted Richard Halford, vice-president at {\color{teal} Old} {\color{blue} Security} in charge of union group insurance, and related the information he had received.  Halford assured Lee that he was aware of Hauser's reputation, but that Hauser was no longer involved with Farmers National.  Halford then called K{\color{violet} avan}agh, who told him that Hauser had no official capacity with the company, and that the financial problems had been cleared up.  Halford did not attempt to check the accuracy of K{\color{violet} avan}agh's representations with the Florida Department of Insurance.

9

Hauser controlled a second company, {\color{cyan} Family} Provider Life Insurance Company ("{\color{cyan} Family} Provider").  In 1975, the company had no business, no office, and {\color{magenta} assets} of \$50,000.  Because of Farmers National's insolvency, Hauser decided to activate

& Family Provider, and its {\color{magenta} assets} were increased to \$250,000, the minimum required to conduct business in Arizona, where the company was licensed.  In January 1976, Boden and K{\color{violet} avan}agh met with Halford and Robert Barton, president of {\color{teal} Old} {\color{blue} Security}, to propose a new agreement between Old Security and {\color{cyan} Family} Provider for the purpose of obtaining the Fund business.  Both Barton and Halford considered Family Provider and Farmers National to be "synonymous" and believed that Kavanagh and Boden

& {\color{blue} Security}

{\color{teal} Old}

{\color{cyan} Family}

{\color{violet} avan}

{\color{magenta} assets}\\
                  \bottomrule
    \end{tabular}}
    \caption{Examples of sequences from in-domain test Pile domains. We highlight the tokens from the evaluation sequence which benefit the most from the summary vectors. In Books3, \emph{L'Humanit\'e} is prominent French newspaper associated with Breton and his circle. In GitHub, the summary vectors carry information about the logical and syntactical continuation of the context.}
    \label{tab:examples_in}
\end{table*}

\begin{table*}[t]
    \centering
    \resizebox{\textwidth}{!}{
    \begin{tabular}{r p{14cm} p{7cm} p{2cm}}
        \toprule
          \textbf{Domain} & \textbf{Compressed context} & \textbf{Evaluation sequence} & \textbf{Most \mbox{improved} \mbox{tokens}}\\
         \midrule
HackerNews 
& Hackers steer {\color{blue} Tesla} into oncoming traffic by placing three stickers on the road - velmu
https://www.businessinsider.com/{\color{blue} tesla}-hackers-steer-into-oncoming-traffic-with-stickers-on-the-road-2019-4

======
chrisbolt
From yesterday:

[https://news.ycombinator.com/item?id=19536375]

(https://news.ycombinator.com/item?id=19536375)

------

gregmac

While I'm hugely skeptical of the current state of self-driving cars, you
could probably get human drivers to make the same mistake if you were to
repaint the lines. However, humans will also notice the oncoming cars (if
there are any) and avoid getting in a head-on collision.

The thing missing from this {\color{teal} test} is that critical practical piece: if there
was an oncoming car, will the {\color{blue} Tesla} do something to avoid the collision? I
would assume that not getting in a head-on crash is higher priority than
staying in the lane {\color{violet} markings}.

Without oncoming traffic, all this is testing is what the {\color{blue} Tesla} considers
valid line {\color{violet} markings}. I'm sure there's room for improvement here (such as
checking where the other lane is, raising the requirement for how well-defined
the lines have to be, etc), but

& those are also going to involve trade-offs
where there are legitimate situations that will stop working.

I think you could just as easily title this video "{\color{blue} Tesla} {\color{magenta} auto}-pilot follows
road {\color{violet} markings} even if they're really bad".

Edit: The best shot I could get from the video [1] makes me even more upset at
this {\color{teal} test}: these look like the temporary markings often used during
construction, just before they come and paint the normal lines using the big

& {\color{blue} Tesla}

{\color{teal} test}

{\color{violet} markings}

{\color{magenta} auto}

\\
ArXiv 
& $z_{{\color{violet} k}} = {\color{cyan} h}_{{\color{violet} k}} \left(x_{{\color{violet} k}}\right) + v_{{\color{violet} k}}, \qquad v_{{\color{violet} k}} \sim \mathcal{N}\left(0,R_{{\color{violet} k}}\right)$

In the above equations, we see that the transition matrix $F_{{\color{violet} k},{\color{violet} k}-1}$ has been replaced by the non{\color{teal} linear} vector-valued function $f_{{\color{violet} k},{\color{violet} k}-1}\left(\cdot\right)$, and similarly, the matrix $H_{\color{violet} k}$, which transforms a vector from the state space into the measurement space, has been replaced by the non{\color{teal} linear} vector-valued function ${\color{cyan} h}_{\color{violet} k}\left(\cdot\right)$. The method proposed by the {\color{blue} Extended} {\color{magenta} Kal}man Filter is to {\color{teal} linear}ize the non{\color{teal} linear}ities about the current state prediction (or estimate). That is, we choose $F_{{\color{violet} k},{\color{violet} k}-1}$ as the Jacobian of $f_{{\color{violet} k},{\color{violet} k}-1}$ evaluated at $\hat{x}_{{\color{violet} k}-1|{\color{violet} k}-1}$, and $H_{\color{violet} k}$ as the Jacobian of ${\color{cyan} h}_{\color{violet} k}$ evaluated at $\hat{x}_{{\color{violet} k}|{\color{violet} k}-1}$ and proceed as in the {\color{teal} linear} {\color{magenta} Kal}man Filter of Section $sec::kf$.[\^18] Numerical accuracy of these methods tends to depend heavily on the non{\color{teal} linear} functions. If we have {\color{teal} linear} constraints but

& a non{\color{teal} linear} $f_{{\color{violet} k},k-1}\left(\cdot\right)$ and ${\color{cyan} h}_k\left(\cdot\right)$, we can adapt the {\color{blue} Extended} {\color{magenta} Kal}man Filter to fit into the framework of the methods described thus far.

Nonlinear Equality and Inequality Constraints
---------------------------------------------

Since equality and inequality constraints we model are often times nonlinear, it is important to make the extension to nonlinear equality and inequality constrained Kalman Fil

& {\color{blue} Extended}

{\color{teal} linear}

{\color{cyan} h}

{\color{violet} k}

{\color{magenta} Kal}\\
Gutenberg 
& eight or nine cents.  Telegrams in foreign languages are
sent within the empire for five sen per word, with a minimum charge of
twenty-five sen for five words or a fraction thereof.  No charge is
made for delivery within a radius of 2-1/2 miles of the telegraph office.

There are no private telegraph corporations.  The government builds,
owns, and operates the lines just as it does the mails.  The {\color{teal} postal} and
{101} telegraph systems are intimately connected, and the same office
does service for both.

The first telegraph line in {\color{cyan} Japan} was opened in 1869.  The venture
proving a success, the following year the line was extended and a
general telegraphic system for the whole country decided upon.  The
rapid construction of telegraph lines began in 1872, from which year it
has gone forward uninterruptedly.  At present the lines extend to every
corner of the empire.  The first lines were surveyed, built, and
operated under foreign experts; but the natives have learned so rapidly
that they have been enabled to do away with all foreign employees.  All
of the materials and instruments in use, with the exception of
submarine cables and the most delicate electrical measuring apparatus,
are made in {\color{cyan} Japan}.

MAILS.--The {\color{cyan} Japanese} {\color{teal} mail} system was modeled after the American in
1871. 
& At first it was {\color{blue} limited} to {\color{teal} postal} service between the three
large cities of {\color{cyan} Tokyo}, Kyoto, and Osaka; but in 1872 it was extended to
the whole country, with the exception of a certain part of the
Hokkaido, which was without roads and almost without population.
To-day there is no village or hamlet in the whole land which does not
enjoy the convenience of a good postal system.  The mails are sent with
promptness and
& {\color{blue} limited}

{\color{teal} postal}

{\color{cyan} Tokyo} \\

YoutubeSubtitles
& te que no voy a esa escuela."

{\color{cyan} Johnny} {\color{blue} Galecki}

El {\color{violet} Dr}. Leonard Hofstadter obtuvo su doctorado
a los 24 años, pero el actor que lo interpreta
sólo llegó a medio camino de la secundaria.
En una entrevista con Time Out Chicago en
el 2009, Johnny Galecki reveló que abandonó
la escuela a mediados del octavo grado luego
de años de evitar ir a clases a toda costa.
Le dijo a Time Out,
"Una vez que las divisiones largas aparecieron
en tercer grado, iba al baño por 45 minutos
y nadie lo notaba, todos los días a la misma
hora del día, sólo para escapar de ellas."
Puede que Galecki no tenga un cerebro matemático,
pero siempre tuvo inteligencia callejera.
"El conocimiento es el mejor y más seguro
tesoro... Vaya, me aburro a mí mismo."
A los 14 años, vivió solo en un apartamentito
en Burbank, California, mient

& ras tr{\color{teal} aba}jaba
en la comedia American Dreamer, su primer
gran trabajo. Su familia pasó nueve meses
en Long Beach antes de regresar a Chicago,
y él se quedó para concentrarse en su carrera
como actor.

{\color{cyan} Jim} {\color{blue} Parsons}

El {\color{violet} Dr}. Sheldon Cooper fue un niño prodigio.
Comenzó la universidad cuando tenía 11 años
& {\color{blue} Parsons}

{\color{teal} aba}

{\color{cyan} Jim}

{\color{violet} Dr}
\\
                  \bottomrule
    \end{tabular}}
    \caption{Examples of sequences from out-of-domain test Pile domains. We highlight the tokens from the evaluation sequence which benefit the most from the summary vectors. In Gutenberg, `Tokyo' is not copied over from the compressed context but is inferred from the discussion of Japan. In YoutubeSubtitles, `Jim Parsons' benefits the most from the summary vectors because the context discusses his co-star John Galecki in \emph{The Big Bang Theory}.}
    \label{tab:examples_out}
\end{table*}

\begin{table*}[t]
    \centering
    \resizebox{\textwidth}{!}{
\begin{tabular}{clccccccc}
    \toprule
    \multirow{2}{*}{\textbf{Dataset}} & \multirow{2}{*}{\textbf{Prompt template}} & \multicolumn{3}{c}{\textbf{OPT-based models}} & \multicolumn{3}{c}{\textbf{Llama-2-based models}} \\
    \cmidrule(lr){3-5} \cmidrule(lr){6-8}
    & & \textbf{Toks. / dem.} & \textbf{Cal.} & \textbf{Bal.} & \textbf{Toks. / dem.} & \textbf{Cal.} & \textbf{Bal.} \\
    \toprule
    AG News & \texttt{Article: \{text\}\textbackslash nTopic: \{label\}} & 65 & \checkmark & & 75 & \checkmark & \\
    \midrule
    SST-2 & \texttt{Sentence: \{sentence\}\textbackslash nSentiment: \{label\}} & 22 & \checkmark & \checkmark & 25 & \checkmark & \checkmark \\
    \midrule
    BoolQ & \texttt{\{passage\}\textbackslash nquestion: \{question\}?\textbackslash nanswer: \{label\}} & 165 & \checkmark & & 170 & \checkmark & \\
    \midrule
    {WiC} & \texttt{\{sentence1\}\textbackslash n\{sentence2\}\textbackslash nquestion: Is the word '\{word\}' } & 45 & \checkmark & & 45 & \checkmark & \\
    & \texttt{used the same way in the two sentences above?\textbackslash nanswer: \{label\}} & & & & & \checkmark & \\
    \midrule
    {WSC} & \texttt{Question: In the sentence "\{text\}", does the pronoun '\{span2\_text\}'} & 61 & & & 50 & \checkmark & \\
    & \texttt{refer to \{span1\_text\}?\textbackslash nAnswer: \{label\}} & & & & &  & \\
    \midrule
    RTE & \texttt{\{premise\}\textbackslash nquestion: \{hypothesis\} True or False?\textbackslash nanswer: \{label\}} & 75 & & & 85 & & \\
    \midrule
    CB & \texttt{\{premise\}\textbackslash nquestion: {hypothesis}. true, false or neither?\textbackslash nanswer: \{label\}} & 98 & & \checkmark & 95 & & \checkmark \\
    
    \midrule
    COPA & \texttt{Context: \{premise\}\textbackslash nAnswer: \{answer\}} & 21 & & \checkmark & 22 & \checkmark & \checkmark \\
    \midrule
    MultiRC & \texttt{Context: \{paragraph\}\textbackslash n\{question\}\textbackslash n\{answer\}\textbackslash nanswer: \{label\}} & 350 & \checkmark & \checkmark & 350 & \checkmark & \checkmark \\
    \midrule
    MR & \texttt{Review: \{text\}\textbackslash nSentiment: \{label\}} & 36 & & \checkmark & 40 & \checkmark & \checkmark \\
    \midrule
    Subj & \texttt{input: \{text\}\textbackslash ntype: \{label\}} & 40 & & \checkmark & 40 & \checkmark & \checkmark \\
    \bottomrule
\end{tabular}
}
        \caption{Details of the datasets and prompts used for the ICL evaluation of our OPT-2.7B and Llama-2-7B AutoCompressors and baselines. ``Toks / dem.'' (\emph{Tokens per demonstration}) denotes how long demonstrations are for the average example. ``Cal.'' (\emph{Calibration}) denotes whether we use calibration \cite{sachan-etal-2022-improving}, and ``Bal.'' (\emph{Balanced}) means whether we enforce class-balanced sampling. We decide the ticks based on which method performs best on a held-out validation set.}
    \label{tab:icl_datasets}
\end{table*}

 \begin{table*}[t]
    \centering
    \resizebox{\linewidth}{!}{\begin{tabular}{rcccccccccccc}
        \toprule
          & \textbf{AG News} & \textbf{SST-2} & \textbf{BoolQ} & \textbf{WiC} & \textbf{WSC} & \textbf{RTE} & \textbf{CB} & \textbf{COPA} & \textbf{MultiRC} & \textbf{MR} & \textbf{Subj}\\
         \bottomrule
         \\
         \multicolumn{12}{c}{OPT-2.7B AutoCompressor} \\
         \toprule
         Zero-shot & $68.2_{(0.0)}$ & $78.0_{(0.0)}$ & $\textbf{60.2}_{(0.0)}$ & $49.5_{(0.0)}$ & $60.6_{(0.0)}$ & $55.2_{(0.0)}$ & $43.6_{(0.0)}$ & $69.0_{(0.0)}$ & $43.8_{(0.0)}$ & $60.0_{(0.0)}$ & $56.7_{(0.0)}$ \\
         50 summary vecs. & ${\textbf{72.7}}_{(1.4)}$ & $84.3_{(9.2)}$ & $55.8_{(4.2)}$ & $50.4_{(1.0)}$ & $61.3_{(5.8)}$ & $54.8_{(3.4)}$ & $55.9_{(5.4)}$ & $71.6_{(0.6)}$ & $44.1_{(1.1)}$ & $70.4_{(10.2)}$ & $\textbf{63.2}_{(7.7)}$\\
        100 summary vecs. & $71.2_{(3.8)}$ & $\textbf{87.0}_{(3.5)}$ & $57.5_{(4.6)}$ & $50.7_{(1.0)}$ & $60.2_{(6.7)}$ & $55.5_{(2.5)}$ & $54.4_{(4.0)}$ & ${\textbf{71.9}}_{(0.4)}$ & $45.6_{(2.8)}$ & ${73.1}_{(12.9)}$ & $62.2_{(5.8)}$\\
        150 summary vecs. & $68.2_{(3.3)}$ & $82.6_{(5.6)}$ & $59.8_{(1.8)}$ & ${\textbf{51.8}}_{(1.1)}$ & ${\textbf{63.5}}_{(0.0)}$ & $\textbf{55.8}_{(1.8)}$ & ${\textbf{58.3}}_{(5.1)}$ & $71.4_{(0.5)}$ & $\textbf{46.7}_{(2.1)}$ & $67.0_{(11.9)}$ & $58.5_{(6.7)}$ \\
        ICL (150 tokens)  & $72.5_{(2.5)}$ & $70.8_{(12.6)}$ & ${60.2}_{(0.0)}$ & $50.4_{(1.1)}$ & $52.3_{(13.9)}$ & $57.6_{(4.3)}$ & $51.1_{(7.1)}$ & $71.3_{(1.5)}$ & $43.8_{(0.0)}$ & $\textbf{86.4}_{(4.2)}$ & $61.7_{(11.2)}$\\
        ICL (750 tokens)  & $67.3_{(3.4)}$ & ${87.5}_{(5.0)}$ & ${69.1}_{(1.0)}$ & $51.0_{(1.7)}$ & $62.9_{(0.8)}$ & ${57.4}_{(4.4)}$ & $49.0_{(1.1)}$ & $72.0_{(0.7)}$ & ${52.0}_{(5.4)}$ & ${86.7}_{(5.9)}$ & ${73.6}_{(13.9)}$ \\
         \bottomrule 
         \\
         \multicolumn{12}{c}{OPT-2.7B RMT} \\
         \toprule
            Zero-shot  & ${66.9}_{(0.0)}$ & $72.8_{(0.0)}$ & ${\textbf{58.4}}_{(0.0)}$ & $50.3_{(0.0)}$ & ${\textbf{64.4}}_{(0.0)}$ & ${55.2}_{(0.0)}$  & $42.2_{(0.0)}$ & ${68.8}_{(0.0)}$ & $43.9_{(0.0)}$ & ${62.5}_{(0.0)}$ & $\textbf{69.8}_{(0.0)}$\\   
            1-step summary vecs. & $66.3_{(5.5)}$ & $86.5_{(5.1)}$ & $49.6_{(8.1)}$ & ${51.0}_{(1.00}$ & $57.7_{(6.6)}$ & $51.3_{(1.2)}$ & ${\textbf{53.3}}_{(3.8)}$ & $67.4_{(1.1)}$ & $44.9_{(1.2)}$ & $52.6_{(2.8)}$ &  $63.3_{(11.2)}$\\
            2-step summary vecs. & $65.2_{(7.2)}$ & ${\textbf{88.6}}_{(2.3)}$ & $54.8_{(4.1)}$ & $50.3_{(0.8)}$ & $58.6_{(6.7)}$ & $50.2_{(1.4)}$ & $49.5_{(4.8)}$ & $68.2_{(1.2)}$ & $\textbf{45.5}_{(1.8)}$ & $54.1_{(1.9)}$ & $54.6_{(1.7)}$\\
            3-step summary vecs. & $63.9_{(3.3)}$ & $84.5_{(6.6)}$ & $41.8_{(9.7)}$ & $50.6_{(0.6)}$ & $54.3_{(7.9)}$  & $50.2_{(1.4)}$ & $49.5_{(3.6)}$ & $68.0_{(0.9)}$ & $45.5_{(1.0)}$ & $52.8_{(1.6)}$ & $58.4_{(8.6)}$\\
            ICL (150 tokens) & ${\textbf{70.8}}_{(1.9)}$  & $75.1_{(13.3)}$ & $58.4_{(0.0)}$ & ${\textbf{51.7}}_{(2.8)}$  & $52.5_{(13.1)}$ & $\textbf{57.2}_{(3.6)}$  & $46.5_{(3.6)}$ & ${\textbf{69.3}}_{(1.5)}$ & $43.9_{(0.0)}$ & ${\textbf{89.0}}_{(1.4)}$ &  $60.7_{(12.1)}$\\
            ICL (750 tokens) & $65.8_{(4.2)}$ & $85.7_{(9.7)}$ & $57.2_{(7.6)}$ & $51.5_{(2.7)}$ & $59.2_{(8.5)}$ & ${57.8}_{(2.0)}$ & $48.2_{(0.7)}$ & $70.9_{(0.7)}$ & ${54.6}_{(3.6)}$ & $87.5_{(4.6)}$ & ${71.6}_{(12.6)}$\\
        \bottomrule
        \\
        \multicolumn{12}{c}{OPT-2.7B Pre-trained} \\
        \toprule
        Zero-shot         & $65.1_{(0.0)}$ & $79.1_{(0.0)}$ & $55.8_{(0.0)}$ & $49.4_{(0.0)}$ & $53.9_{(0.0)}$ & $51.2_{(0.0)}$ & $21.2_{(0.0)}$ & $66.8_{(0.0)}$ & $43.7_{(0.0)}$ & $59.0_{(0.0)}$ & $66.2_{(0.0)}$\\
        ICL (150 tokens)  & $71.6_{(2.6)}$ & $68.56_{(14.9)}$& $55.8_{(0.0)}$ & $50.6_{(1.0)}$ & $53.30_{(11.1)}$& $56.1_{(2.4)}$ & $46.2_{(6.4)}$ & $71.7_{(1.2)}$ & $43.7_{(0.0)}$ & $86.7_{(4.3)}$ &  $61.9_{(10.9)}$\\
        ICL (750 tokens)  & $63.3_{(5.1)}$  & $91.0_{(3.2)}$  & $63.0_{(1.3)}$  & $50.0_{(0.4)}$  & $63.5_{(0.6)}$  & $54.7_{(3.0)}$  & $52.1_{(4.8)}$  & $73.4_{(1.0)}$  & $53.5_{(6.2)}$ & $89.9_{(2.2)}$ & $64.4_{(10.7)}$\\
        \bottomrule
         \\
         \multicolumn{12}{c}{Llama-2-7B AutoCompressor} \\
         \toprule
        Zero-shot         & $63.3_{(0.0)}$ & $67.7_{(0.0)}$ & $67.4_{(0.0)}$ & $50.8_{(0.0)}$ & $43.3_{(0.0)}$ & $58.8_{(0.0)}$ & $42.9_{(0.0)}$ & $52.5_{(0.0)}$ & $52.5_{(0.0)}$ & $57.4_{(0.0)}$ & $49.3_{(0.0)}$ \\
        50 summary vecs.  & $79.6_{(4.9)}$ & ${\textbf{94.2}}_{(1.6)}$ & ${\textbf{70.1}}_{(3.3)}$ & $51.6_{(2.1)}$ & $47.7_{(8.7)}$ & $66.3_{(7.0)}$ & $46.4_{(18.7)}$ & $84.5_{(1.0)}$ & $52.6_{(2.8)}$ & $\textbf{91.5}_{(1.0)}$ & $53.5_{(3.6)}$ \\
        100 summary vecs. & ${\textbf{87.6}}_{(1.2)}$ & $92.6_{(3.3)}$ & $66.3_{(2.8)}$ & $52.5_{(2.2)}$ & $42.9_{(2.5)}$ & $63.5_{(6.6)}$ & ${\textbf{64.5}}_{(5.9)}$ & $85.9_{(0.4)}$ & ${\textbf{56.1}}_{(1.2)}$ & $90.7_{(2.6)}$ & $\textbf{57.0}_{(5.6)}$ \\
        150 summary vecs. & $85.4_{(3.4)}$ & $92.3_{(2.9)}$ & $68.0_{(1.8)}$ & ${\textbf{52.8}}_{(1.5)}$ & $49.9_{(7.6)}$ & $65.3_{(6.6)}$ & $54.8_{(5.8)}$ & ${\textbf{86.1}}_{(0.6)}$ & $54.8_{(2.2)}$ & $91.1_{(2.2)}$ & $56.6_{(7.9)}$ \\
        ICL (150 tokens)  & $74.5_{(2.2)}$ & $92.4_{(3.1)}$ & $67.4_{(0.0)}$ & $52.4_{(2.7)}$ & ${\textbf{51.8}}_{(6.9)}$ & $\textbf{69.1}_{(2.1)}$ & $46.4_{(23.0)}$ & $80.0_{(1.9)}$ & $52.5_{(0.0)}$ & $79.7_{(15.7)}$ & $57.9_{(10.7)}$ \\
        ICL (750 tokens)  & $81.2_{(4.1)}$ & $93.8_{(1.2)}$ & $67.7_{(2.7)}$ & $52.4_{(2.0)}$ & $40.0_{(5.7)}$ & ${73.1}_{(3.5)}$ & $50.3_{(2.8)}$ & $82.6_{(1.6)}$ & $47.0_{(3.2)}$ & ${91.6}_{(0.8)}$ & ${{60.7}}_{(14.8)}$ \\
        \bottomrule
        \\
        \multicolumn{12}{c}{Llama-2-7B Pre-trained} \\
        \toprule
        Zero-shot         & $68.8_{(0.0)}$ & $87.2_{(0.0)}$ & $70.0_{(0.0)}$ & $51.4_{(0.0)}$ & $65.4_{(0.0)}$ & $62.8_{(0.0)}$ & $32.1_{(0.0)}$ & $75.5_{(0.0)}$ & $54.5_{(0.0)}$ & $84.1_{(0.0)}$ & $48.9_{(0.0)}$ \\
        ICL (150 tokens)  & $71.9_{(3.8)}$ & $91.6_{(2.9)}$ & $70.0_{(0.0)}$ & $51.0_{(1.9)}$ & $55.4_{(3.2)}$ & $70.9_{(1.7)}$ & $39.3_{(21.2)}$ & $84.2_{(1.3)}$ & $54.5_{(0.0)}$ & $90.6_{(3.3)}$ & $63.6_{(10.8)}$ \\
        ICL (750 tokens)  & $78.2_{(3.8)}$ & $94.5_{(0.8)}$ & $70.3_{(6.1)}$ & $54.9_{(1.9)}$ & $42.2_{(5.0)}$ & $71.3_{(4.4)}$ & $51.3_{(3.5)}$ & $85.3_{(0.7)}$ & $47.0_{(1.5)}$ & $92.9_{(0.5)}$ & $65.4_{(14.5)}$ \\
         \bottomrule

    \end{tabular}}
        \caption{We evaluate the following models on 11 in-context learning tasks: The OPT-2.7B AutoCompressor and RMT model, the Llama-2-7B  AutoCompressor, and the respective pre-trained models. For each fine-tuned model, numbers in bold are the highest evaluation results using at most 150 additional tokens. When using summary vectors, the OPT-2.7B AutoCompressor outperforms the RMT model on 8/11 tasks. Moreover, the OPT-2.7B AutoCompressor benefits from multiple compression steps on most tasks whereas the RMT model performs best without summary vectors on 7/11 tasks and benefits from 3-step summary vectors on none of the above tasks.      The Llama-2 AutoCompressor achieves the absolute highest accuracy using summary vectors on 7/11 tasks. It also achieves the highest accuracy with summary vectors on 9/11 tasks using at most 150 additional tokens.}
    \label{tab:expanded_icl_results}
\end{table*}

\end{document}